\documentclass{article}
\usepackage{iclr2027_conference,times}

\usepackage{graphicx}
\usepackage{wrapfig}
\usepackage{placeins}
\usepackage{needspace}
\usepackage{amsmath}
\usepackage{amssymb}
\usepackage{booktabs}
\usepackage{multirow}

\usepackage{xcolor}
\usepackage{hyperref}
\usepackage{url}
\usepackage[capitalize,noabbrev]{cleveref}

\definecolor{cvprblue}{rgb}{0.21,0.49,0.74}
\hypersetup{colorlinks,urlcolor=black,citecolor=cvprblue}

\title{WorldExpand: Building Coherent 3D Worlds via Sparse Joint Panoramic Expansion}

\author{
\bfseries
Yuyang Yin$^{1,2}$\enspace Chenjie Cao$^{2,\dagger}$\enspace Xuhui Zuo$^{2}$\enspace Yuning Gong$^{2}$\enspace Pengfei Wang$^{2}$ \\
\bfseries
Junta Wu$^{2}$\enspace Zhenwei Wang$^{2}$\enspace  Yao Zhao$^{1}$\enspace Tengfei Wang$^{2}$\thanks{Corresponding authors.\qquad $^{\dagger}$Project leader.}\enspace Yunchao Wei$^{1}$\footnotemark[\value{footnote}]\enspace Chunchao Guo$^{2}$\footnotemark[\value{footnote}] \\
\mdseries
$^{1}$Beijing Jiaotong University \quad $^{2}$Tencent Hunyuan
}

\iclrfinalcopy
\begin{document}

\maketitle
\lhead{Preprint}

\vspace{-0.2in}
\begin{figure}[h]
\centering
\includegraphics[width=\linewidth]{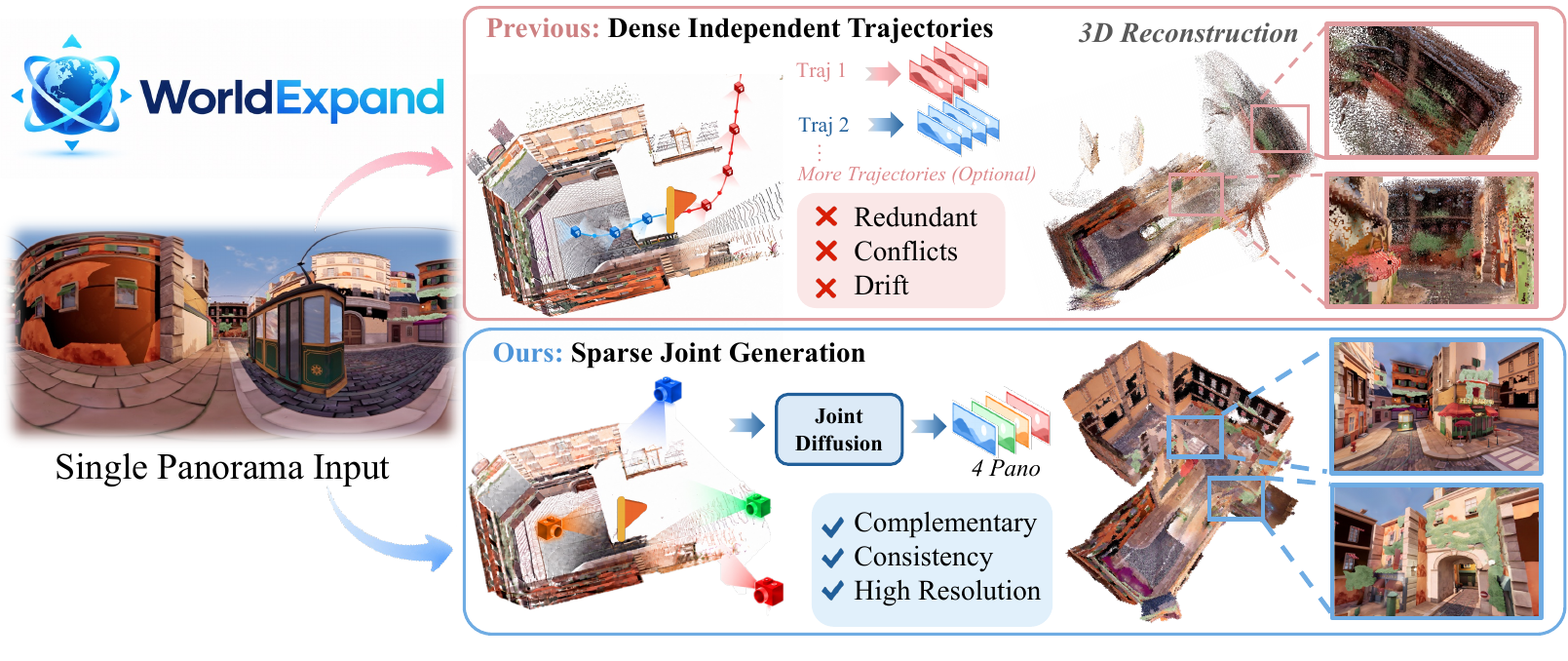}
\vspace{-0.3in}
\caption{\textbf{WorldExpand: Sparse joint panorama generation for 3D world expansion.} Given a single panorama, previous panoramic-video methods~\citep{yang2025matrix} rely on independently generated dense trajectories to explore multiple directions. These trajectories contain redundant observations, may conflict in overlapping regions, and accumulate drift, making them unreliable for 3D reconstruction. In contrast, WorldExpand selects sparse, complementary viewpoints and jointly synthesizes four consistent, high-resolution panoramas, enabling more complete and detailed 3D reconstruction for world exploration.\label{fig:teaser}}
\vspace{-0.15in}
\end{figure}

\begingroup
\setlength{\leftmargini}{2.5pc}
\begin{abstract}
Constructing complete 3D worlds from a single panorama is challenging because large portions of the scene remain unobserved. Existing panoramic-video methods expand spatial coverage through dense sequential generation, but neighboring frames contain substantial redundancy, long trajectories accumulate errors, and independently generated directions may be inconsistent.
We present \textbf{WorldExpand}, which formulates single-panorama world expansion as the generation of sparse, complementary, and high-resolution panoramic observations. This representation concentrates the generation budget on informative scene regions and preserves the fine-grained details required for 3D construction. WorldExpand addresses two key challenges: \textbf{where to observe} and \textbf{how to synthesize}. Our \emph{Coverage-Guided View Selection} identifies informative target locations using scene-adaptive quadrants and warp-based coverage scores, while our \emph{Joint Panoramic Diffusion} uses Gaussian-splatting-based geometric warps to jointly synthesize four pose-grounded and cross-view-consistent panoramas. 
A diagnostic-guided global--local attention schedule retains global attention only in sensitive layers, reducing the training and inference costs of high-resolution joint generation. The resulting posed panoramas can be directly fed into an existing reconstruction pipeline to build an explicit 3D representation.
Experiments show that WorldExpand recovers content behind occluders and around distant corners, improves both novel-view synthesis and downstream 3D reconstruction metrics across various datasets, and generalizes robustly to diverse in-the-wild scenes. WorldExpand runs significantly faster than the state-of-the-art panoramic-video method while producing higher-resolution observations with stronger cross-view consistency.
{\hypersetup{urlcolor=cvprblue}\mbox{\url{https://yuyangyin.github.io/WorldExpand_web/}}}
\end{abstract}
\endgroup

\section{Introduction}

Constructing explorable and geometrically coherent 3D worlds from text or a single image is a long-standing goal with broad applications in immersive content creation and virtual production~\citep{hunyuanworld2025tencent,hy2026hy,lu2025genex,zhou2024dreamscene360}, games and embodied simulation~\citep{gan2020threedworld,savva2019habitat}, and spatial intelligence~\citep{wu2026spatial,liu2025spatialcot,liu2026spatial}. Video-based novel-view synthesis (NVS) methods extrapolate perspective observations from limited inputs~\citep{zhang2026worldstereo,yu2024viewcrafter,cao2025uni3c,wan2025wan,kong2024hunyuanvideo}, but their narrow fields of view require multiple directional clips for full angular coverage. An initial panorama---captured directly or synthesized from text or an image by existing panorama generators~\citep{panfusion2024,feng2025dit360,hy2026hy}---provides full $360^\circ$ angular coverage in the image domain, where scalable generative priors are readily available, but remains spatially incomplete: its single viewpoint cannot reveal surfaces behind occluders, around corners, or beyond doorways.

Expanding beyond this fixed viewpoint requires moving the observation center and synthesizing panoramas at new locations. Panoramic video generation~\citep{lu2025genex,yin2025panoworld,wang2024360dvd,yang2025matrix} offers a natural route, but existing approaches inherit fundamental limitations from dense sequential generation. In particular, they suffer from two limitations. (i)~\emph{Redundant information representation}: densely sampled consecutive frames encode largely overlapping scene information and thus provide redundant multi-view evidence for reconstruction. (ii)~\emph{Limited and inconsistent exploration}: a single rollout explores primarily along one direction, independently generated rollouts toward multiple directions may disagree in overlapping regions, and errors accumulate over long trajectories, resulting in inconsistent multi-view observations for 3D reconstruction. Consequently, directly adopting the output of video generation as the intermediate representation is not well aligned with the goal of 3D world construction. This raises a fundamental question: \textbf{What representation can capture maximal information at minimal generation cost while being tailored to 3D world construction?}

To address this question, we introduce \textbf{WorldExpand}, a framework that formulates single-panorama world expansion as the generation of a sparse set of high-resolution panoramic observations at novel locations. Each panorama captures the full angular context at one location, while a few displaced panoramas provide complementary spatial coverage. Compared with dense video, this representation avoids spending computation on redundant neighboring frames and instead uses the saved budget to increase observation resolution, which is crucial for preserving the fine-grained structures required for detailed 3D reconstruction.

Given this sparse panoramic representation, two key challenges remain. (1)~\textbf{Where should the new observations be placed?} Fixed target directions ignore scene layout and may waste the limited view budget on blocked or redundant regions. Our \emph{Coverage-Guided View Selection} adapts the quadrant orientation to scene boundaries and selects an informative target in each region. (2)~\textbf{How can we jointly generate consistent high-resolution panoramas?} Our \emph{Joint Panoramic Diffusion} uses Gaussian-splatting warps as geometric anchors and jointly synthesizes all four target panoramas to maintain cross-view consistency. To further eliminate unnecessary computation, we introduce a diagnostic-guided global--local attention schedule. Our diagnostic reveals that not all layers require global attention: sensitive layers retain cross-view communication, while the remaining layers use local attention, substantially reducing the training and inference costs of high-resolution joint generation. The resulting posed panoramas can be directly consumed by an existing 3D reconstruction pipeline~\citep{hy2026hy} to build the expanded world.

Experiments show that WorldExpand effectively expands scenes by generating content behind occluders and around distant corners. Across various datasets, it outperforms prior panoramic-video methods on both novel-view synthesis and downstream 3D reconstruction metrics, while generalizing robustly to diverse in-the-wild scenes. Owing to its efficient sparse representation, WorldExpand achieves up to an $8.6\times$ speed-up over the state-of-the-art method~\citep{yang2025matrix}, while generating higher-resolution observations with stronger cross-view consistency. In practical applications, WorldExpand provides a useful building block for large-scale roaming, games, virtual environments, and embodied AI simulation.

\section{Related Work}
\paragraph{Novel-View Synthesis.}
Diffusion-based methods~\citep{liu2023zero,hoellein2024viewdiff,shi2024mvdream,yin20234dgen,liang2024diffusion4d}
hallucinate novel views via pose conditioning or multi-view denoising, while
camera-controllable video diffusion
models~\citep{voleti2024sv3d,wang2024motionctrl,he2024cameractrl,sun2025worldplay,gao2024cat3d,wang2026worldcompass,wu2025cat4d,li2026flashworld,li2026martian}
generate view-consistent sequences along specified trajectories. Other methods
use 3D geometry to guide the synthesis or editing of related
views~\citep{yu2024viewcrafter,huang2025voyager,cao2025uni3c,zhang2026worldstereo,xing2026stereoworld,wang2026edit},
motivating our use of point-cloud warping and joint multi-view denoising.
For static reconstruction, dense narrow-FoV sequences contain redundant observations and can accumulate errors. We instead generate a compact set of strategically placed
\emph{full-view ($360^\circ$) panoramas} that provides complementary coverage
and high-resolution observations for downstream reconstruction.

\paragraph{Panorama Generation.}
Panoramic image generation methods adapt 2D diffusion priors to the
$360^\circ$ domain via panorama–perspective
coupling~\citep{panfusion2024,feng2025dit360,hy2026hy}, cube-face
synchronization~\citep{huang2025dreamcube,kalischek2025cubediff},
sphere-aware attention~\citep{Ye2024DiffPano}, or fused overlapping diffusion
paths~\citep{bar2023multidiffusion}. Video diffusion models also adapt
perspective priors to generate panoramic sequences~\citep{dong2025panolora,xie2025videopanda,fang2026panoramic}.
Panoramic video generation either lifts a
perspective video to $360^\circ$
playback~\citep{li2026cubecomposer,luo2025beyond,wu2026360anything,tan2026imagine360}
or synthesizes panoramic video from text or images for explorable scenes and
long-horizon
roaming~\citep{wang2024360dvd,xing2025tip4gen,lu2025genex,yang2025matrix,yin2025panoworld,li20254k4dgen,xia2025panowan,liu2026omniroam}.
However, image-level methods produce only a single panorama from text or
layout, while long video sequences accumulate appearance drift and geometric
inconsistency, generate overlapping regions across roaming directions
independently, and lack a co-designed reconstruction component. We instead
jointly synthesize a compact set of mutually consistent \emph{novel-view
panoramas} that are simultaneously faithful to a source panorama and to
specified target camera poses, and feed them into a tightly integrated
reconstruction module.

\paragraph{3D Scene Generation.}
Recent 3D scene generation systems differ in the visual observations used for reconstruction.
Panorama-centric methods generate a single $360^\circ$
panorama or world proxy, providing full angular context but remaining tied to
one observation center even when hidden depth layers are completed~\citep{zhou2024dreamscene360,hunyuanworld2025tencent,yang2025layerpano3d}.
To expand coverage, other methods generate camera-controlled videos or
iterative perspective views before fusing them into
3D~\citep{schneider2025worldexplorer,hy2026hy,ainetter2026scene,chung2023luciddreamer,erkocc2026worldagents,yu2024wonderjourney,yu2025wonderworld}.
Object-centric methods also assemble scenes from generated or retrieved assets~\citep{yin2026stateflow,yang2026flowscene}.
However, covering a scene with view-based methods requires generating a large
number of video frames or perspective images, resulting in a redundant and
inefficient intermediate representation. In contrast, WorldExpand uses a
compact set of sparse panoramic observations: it selects complementary
locations and jointly generates their high-resolution panoramas, retaining full
angular context at each location while enforcing cross-view consistency for
downstream reconstruction.

\section{WorldExpand}

\begin{figure*}
    \begin{center}
        \includegraphics[width=1.0\linewidth]{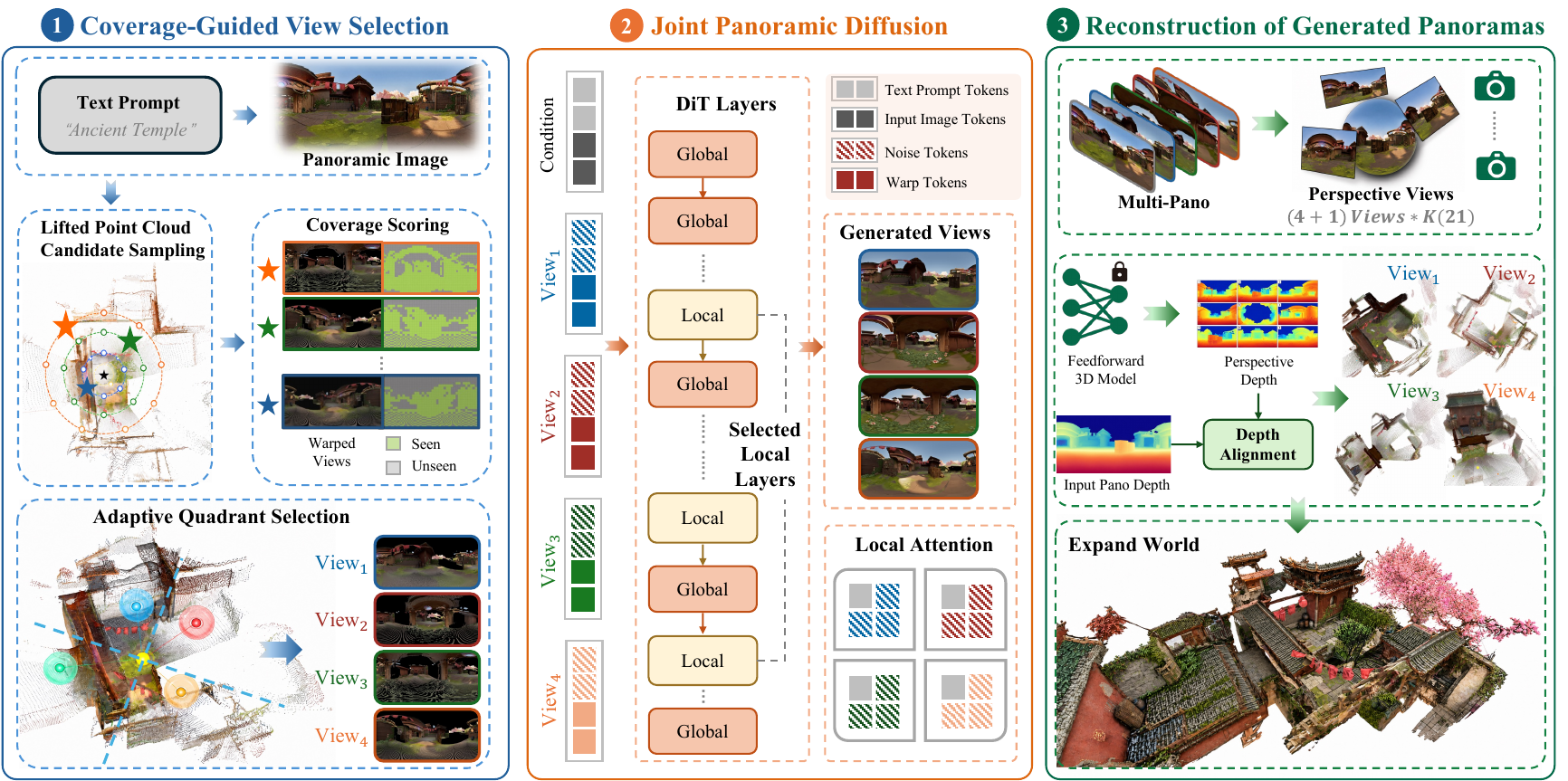}
        \vspace{-0.15in}
        \caption{\textbf{Overview of WorldExpand.} Given one panorama, we select four viewpoints that expose complementary unseen regions, jointly synthesize their panoramic observations with selective cross-view communication, and align multi-view depths to reconstruct an expanded Gaussian scene. \label{fig:method}}
        \vspace{-0.15in}
    \end{center}
\end{figure*}

As discussed above, dense panoramic video is an inefficient representation for 3D world construction: neighboring frames contain largely redundant observations, while long or independently generated trajectories can introduce cross-view inconsistencies. Given a single panorama $I_0$ and its depth $D_0$,  estimated with DepthMaster~\citep{wang2026depthmaster} for in-the-wild inputs, WorldExpand instead expands it into a coherent 3D world through a sparse panoramic representation designed around the needs of reconstruction (\Cref{fig:method}). It first selects a limited set of informative viewpoints to maximize complementary scene coverage (\Cref{sec:view_planning}), then jointly synthesizes their panoramas while maintaining pose grounding and cross-view consistency (\Cref{sec:nvs_generation}), and finally fuses the resulting observations into a coherent 3D representation (\Cref{sec:panorama_aware_reconstruction}).

\subsection{Coverage-Guided View Selection}
\label{sec:view_planning}

\noindent\textbf{Why Coverage-Guided Selection?}
At inference time, we back-project the source panorama $I_0$ and its estimated depth $D_0$ into a point cloud $\mathcal{P}_0$. Although $I_0$ covers all viewing directions, its single observation center leaves large regions occluded. Random or naive target placement ignores scene layout and source visibility, and may therefore select blocked, redundant, or highly overlapping views that fail to reveal these unobserved regions. Given a limited budget of $N{=}4$ target views, we instead select spatially distributed poses according to their expected coverage of newly exposed content.

\noindent\textbf{Candidate Sampling and Coverage Scoring.}
We extract the scene boundary from a horizontal mesh slice at camera height and sample candidate cameras on concentric rings around the source. Candidates that collide with scene geometry or lie beyond the boundary are removed. We then warp $\mathcal{P}_0$ into each candidate view $v$ using Gaussian splatting. Pixels receiving projected source points form the seen region $\mathcal{S}_v$, whereas pixels receiving no points form the unseen region $\mathcal{U}_v$. The coverage score $c_v=\left|\mathcal{U}_v\right|$ measures how much new content candidate $v$ can reveal and is used for the adaptive quadrant selection below.

\noindent\textbf{Adaptive Quadrant Selection.}
To prevent the four targets from clustering in one open direction, we partition the azimuth into four $90^\circ$ quadrants and select one target per quadrant. Since fixed axis-aligned quadrants ignore scene layout, we rotate the dividing cross to align its boundaries with the nearest walls:
\begin{equation}
\label{eq:quadrant_rotation}
\theta^{\star}=\arg\min_{\theta\in[0,\pi/2)}\ \sum_{k=0}^{3} d_{\mathrm{bnd}}\!\left(\theta+k\tfrac{\pi}{2}\right),
\end{equation}
where $d_{\mathrm{bnd}}(\phi)$ is the distance from the source to the scene boundary along azimuth $\phi$. This places low-information wall directions on quadrant boundaries and open regions inside the quadrants. From the candidates $Q_i(\theta^\star)$ in each quadrant, we choose
\begin{equation}
\label{eq:quadrant_selection}
v_i^{\star}=\arg\max_{v\in Q_i(\theta^\star)}c_v,\qquad i=1,\ldots,4.
\end{equation}
Each selected candidate $v_i^\star$ specifies a target camera pose $\mathbf{T}_i$. Relative to the source pose $\mathbf{T}_0$, these target poses define the transformations $\{\Delta\mathbf{T}_{0\rightarrow i}\}_{i=1}^{N}$ used by the next stage to render full-resolution geometric conditions (\Cref{sec:nvs_generation}).

\subsection{Joint Panoramic Diffusion}
\label{sec:nvs_generation}

Given the source panorama $I_0$ and the selected target relative poses $\{\Delta\mathbf{T}_{0\rightarrow i}\}_{i=1}^{N}$ from \Cref{sec:view_planning}, our goal is to jointly synthesize the corresponding novel-view panoramas $\{\hat{I}_i\}_{i=1}^{N}$ using a flow-based image diffusion model~\citep{wu2025qwen} as the generative backbone.
Our main designs are geometry-aware multi-stream conditioning for precise pose grounding and a diagnostic-guided global--local attention schedule for efficient cross-view consistency.

\noindent\textbf{Geometry-Aware Multi-Stream Conditioning.}
Text- or pose-only conditioning lacks spatial anchors, depth-as-color maps are sensitive to distribution shifts, and mesh warping stretches surfaces across depth discontinuities. We instead adopt Gaussian splatting to render geometric warps. Leveraging the accurate poses and depth provided by our synthetic dataset (Supplementary \Cref{app:dataset}), we unproject the source RGB-D panorama $I_0$ and $D_0$ into a colored point cloud $\mathcal{P}_0$, treat each point as an isotropic Gaussian splat, and render the resulting representation from each target pose:
\begin{equation}
    (W_i,M_i)=\mathrm{Render}_{\mathrm{ERP}}(\Delta\mathbf{T}_{0\rightarrow i}\;\mathcal{P}_0),\quad i=1,\ldots,N.
\end{equation}
The warped image $W_i$ provides pixel-level correspondences from the source panorama to the target view, while the mask $M_i$ marks disoccluded regions that lack source observations and must be hallucinated. This conditioning directly translates the pose specification into a spatially explicit generation guide: visible regions are anchored by the warp, and the model's generative capacity is focused on the masked areas. In our $N$-view generator ($N{=}4$ in our experiments), we spatially concatenate all image tokens into one global sequence $\mathbf{H}^{global}$, which contains three semantic streams:
\begin{equation}
\mathbf{H}^{global}=\left[\mathbf{P}_{txt};\mathbf{S}_0;\mathbf{Z}_1;\ldots;\mathbf{Z}_N;\mathbf{W}_1;\ldots;\mathbf{W}_N\right],
\end{equation}
where $\mathbf{P}_{txt}$ indicates textual tokens while $\mathbf{S}_0$ is the source panorama token sequence; $\{\mathbf{Z}_i\}_{i=1}^{N}$ are noisy target-view tokens; $\{\mathbf{W}_i\}_{i=1}^{N}=\{\mathrm{VAE}(W_i)+\phi(M_i)\}_{i=1}^N$ denote warped tokens, formed by channel-wise addition of the VAE-encoded warped image and a convolutional mask feature $\phi(M_i)$ that encodes per-pixel visibility information. To bridge the gap between clean training warps and estimated inference geometry, we independently perturb each target-view warp during training.

\noindent\textbf{Diagnostic-Guided Global--Local Attention Schedule.}
Our global attention over $\mathbf{H}^{global}$ promotes cross-view consistency, but using it at every layer incurs unnecessary computation that grows quadratically with the number of views, especially at high resolution. Since layers serve distinct roles---some align views while others refine local texture---we retain global attention only where cross-view communication is essential and use efficient per-view local attention $\mathbf{H}^{local}_i{=}[\mathbf{P}_{txt};\mathbf{Z}_i]$ elsewhere.

\Needspace{20\baselineskip}
\begin{wrapfigure}{r}{0.55\linewidth}
    \centering
    \vspace{-0.15in}
    \includegraphics[width=\linewidth]{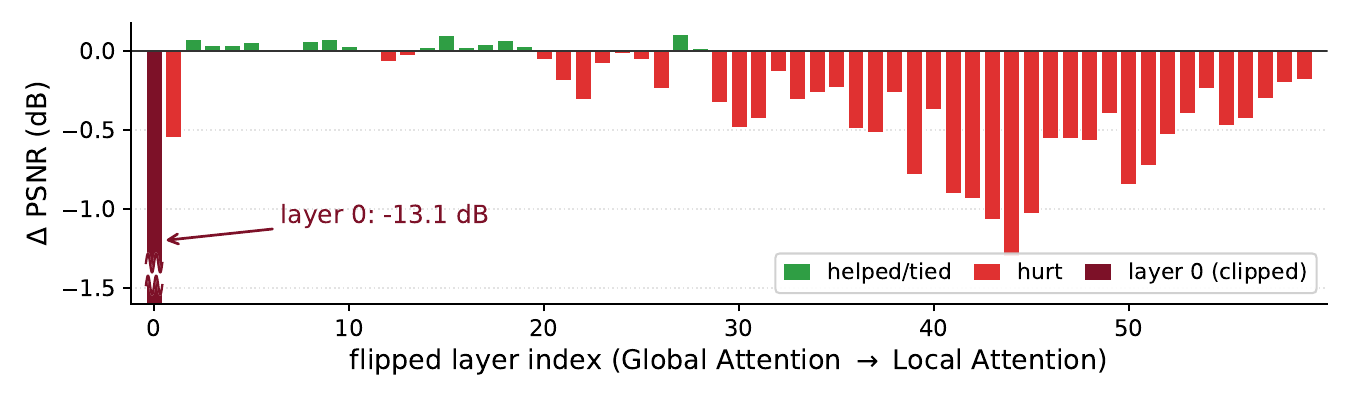}
    \caption{\textbf{Per-layer diagnostic for the global--local attention schedule.} Without finetuning, each bar reports the $\Delta$PSNR when one transformer layer is switched from global to local attention at inference. We use these sensitivity scores to determine which 30 layers are switched to local attention, prioritizing the least-sensitive ones.\label{fig:per_layer_delta_psnr}}
    \vspace{-0.15in}
\end{wrapfigure}

To resolve this, we perform a per-layer diagnostic on a fully global-attention model: each layer is individually switched to local attention without finetuning, and the $\Delta$PSNR is measured against the full-global reference. As shown in \Cref{fig:per_layer_delta_psnr}, the sensitivity is clearly heterogeneous---localizing certain layers causes severe degradation while many others incur negligible loss. We select the 30 layers with minimal degradation as local layers and finetune the resulting model, achieving comparable quality to the full-global baseline with significantly reduced memory and latency.

\subsection{Reconstruction of Generated Panoramas}
\label{sec:panorama_aware_reconstruction}

We reconstruct the scene using the WorldMirror 2.0 and 3DGS
composition pipeline of HY-World 2.0~\citep{hy2026hy}, converting the source
and generated posed panoramas into a unified Gaussian representation. Complete
details of perspective decomposition, depth prediction and alignment, Gaussian
initialization, and optimization are provided in Supplementary
\Cref{sec:reconstruction_details}.

\begingroup
\setlength{\textfloatsep}{16pt plus 2pt minus 2pt}
\section{Experiments}

\subsection{Datasets}

We evaluate on two complementary panoramic RGB-D benchmarks and a diverse set of in-the-wild panoramas. The RGB-D benchmarks comprise our UE dataset for measuring in-domain fidelity and the synthetic subset of RealSee3D for evaluating cross-dataset generalization. 
On each benchmark, we sample $64$ anchor test cases and expand every anchor into $4$ target views, yielding $256$ generated panoramas for image- and geometry-level quantitative evaluation. We further evaluate on diverse in-the-wild panoramas to assess practical generalization when ground-truth depth is unavailable; in this setting, the source depth is estimated using DepthMaster~\citep{wang2026depthmaster}.

\noindent\textbf{UE Dataset.}
Our primary benchmark is a synthetic panoramic RGB-D dataset built in Unreal Engine (Supplementary \Cref{app:dataset}). It contains diverse indoor and outdoor environments with accurate camera poses and dense metric depth, supporting both point-cloud warp conditioning and ground-truth 3D evaluation.

\noindent\textbf{RealSee3D.}
To assess generalization to a distinct scene distribution, we use the synthetic subset of the public RealSee3D benchmark~\citep{Li2025realsee3d_data}, which provides indoor multi-view panoramic RGB-D data with rendered camera poses and metric depth. This benchmark is held out from training and differs from our Unreal Engine training scenes.

\subsection{Evaluation Metrics}
We evaluate along two axes. For \textbf{per-view image quality}, we report PSNR, SSIM, and LPIPS against the ground-truth panorama at each target pose. For \textbf{3D reconstruction quality}, we measure (i)~\emph{camera-pose accuracy}: the poses of the four target panoramas are estimated with the off-the-shelf VGGT-$\Omega$~\citep{wang2026vggt} (each panorama sliced into perspective views) and compared to the ground truth via pose AUC at $3^\circ$/$30^\circ$; and (ii)~\emph{point-cloud geometry}: for every method we reconstruct the point cloud with the \emph{same} Panorama-Aware 3D Reconstruction module (\Cref{sec:panorama_aware_reconstruction}) to ensure a fair comparison, and compare it to the ground-truth point cloud---obtained by unprojecting and merging the ground-truth RGB-D of the source (center) view and the four novel views---reporting Accuracy and Completeness (per-scene mean/median) and F-score@5cm.
Full training and reconstruction implementation details are provided in Supplementary \Cref{sec:implementation_details}.

\subsection{Experimental Results}

To our knowledge, no prior work shares the same setting of expanding a scene from a single panorama via multi-view panoramas. We therefore compare against the closest line of work---panoramic video generation---and select GenEx~\citep{lu2025genex} and Matrix-3D~\citep{yang2025matrix} as representative baselines. Notably, both GenEx and Matrix-3D are trained on synthetic scenes rendered in Unreal Engine, so evaluation on held-out UE scenes does not give our method exclusive access to the synthetic-engine domain. Since our multi-view panoramas may correspond to target locations along arbitrary directions, requiring a single panoramic video to cover all of them is unrealistic. For a fair comparison, we instead run each baseline four times, generating panoramic videos along the four target directions independently, and extract the keyframes corresponding to our target viewpoints for evaluation against the ground truth.

\noindent\textbf{Comparison on UE Synthetic.}
On our in-domain UE synthetic dataset, our method outperforms both baselines on all NVS metrics (\Cref{tab:main_results}); the complete per-view results in Supplementary \Cref{tab:per_view_results_full} further show that this advantage holds across every target view. The PSNR/SSIM gains show our panoramas stay pixel- and structure-level faithful to the source-grounded reference, while the lower LPIPS reflects fewer hallucinated, cross-run-inconsistent structures that would otherwise corrupt downstream 3D reconstruction. This advantage carries over to 3D: because the baselines drift from the source geometry and disagree across independently generated views, both their estimated poses and reconstructed point clouds degrade, whereas our consistency-aware panoramas remain source-grounded and cross-view coherent.

\noindent\textbf{Comparison on RealSee3D.}
We further evaluate on the synthetic subset of the public RealSee3D benchmark to probe cross-dataset generalization. Crucially, \emph{none} of the compared models---neither the baselines nor ours---is trained or fine-tuned on RealSee3D; every method is applied in a purely zero-shot manner. This makes RealSee3D a fair, held-out testbed that measures generalization ability alone, isolating it from any advantage of in-domain training. Under the same protocol, the NVS and 3D reconstruction metrics in \Cref{tab:main_results} follow the same trend as on our UE synthetic benchmark: our method improves both image fidelity and reconstructed geometry, showing that the consistency-aware design transfers beyond the training distribution. Additional qualitative comparisons are provided in Supplementary \Cref{app:realsee}.

\begin{table*}[t]
\centering
\caption{\textbf{Novel-view synthesis and 3D reconstruction.} NVS metrics are averaged over the four target views. AUC and F@5\,cm are percentages; Accuracy and Completeness are reported in meters as mean/median. Best per dataset is in \textbf{bold}.}
\label{tab:main_results}
\scriptsize
\setlength{\tabcolsep}{2.5pt}
\begin{tabular}{ll ccc ccccc}
\toprule
& & \multicolumn{3}{c}{Novel-view synthesis} & \multicolumn{5}{c}{3D reconstruction} \\
\cmidrule(lr){3-5}\cmidrule(lr){6-10}
Dataset & Method & PSNR $\uparrow$ & SSIM $\uparrow$ & LPIPS $\downarrow$ & AUC@$3^\circ\uparrow$ & AUC@$30^\circ\uparrow$ & Acc.$\downarrow$ & Comp.$\downarrow$ & F@5\,cm$\uparrow$ \\
\midrule
\multirow{3}{*}{UE synthetic}
& GenEx    & 15.18 & 0.461 & 0.468 & 4.58  & 20.96 & 0.86/0.46 & 3.70/0.55 & 12.95 \\
& Matrix-3D & 17.30 & 0.502 & 0.363 & 29.38 & 83.96 & 0.80/0.34 & \textbf{2.50}/0.24 & 20.32 \\
& Ours     & \textbf{20.16} & \textbf{0.658} & \textbf{0.256} & \textbf{49.79} & \textbf{89.88} & \textbf{0.79/0.19} & 2.87/\textbf{0.09} & \textbf{29.02} \\
\midrule
\multirow{3}{*}{RealSee3D}
& GenEx    & 15.85 & 0.644 & 0.455 & 0.42 & 19.04 & 0.21/0.14 & 0.31/\textbf{0.15} & 24.42 \\
& Matrix-3D & 17.44 & 0.651 & 0.391 & 8.33 & 67.46 & 0.12/0.08 & \textbf{0.30}/0.17 & 35.04 \\
& Ours     & \textbf{18.06} & \textbf{0.711} & \textbf{0.350} & \textbf{25.00} & \textbf{81.42} & \textbf{0.10/0.06} & \textbf{0.30}/0.17 & \textbf{41.44} \\
\bottomrule
\end{tabular}
\end{table*}

\begin{figure*}
    \centering
    \includegraphics[width=1.0\linewidth]{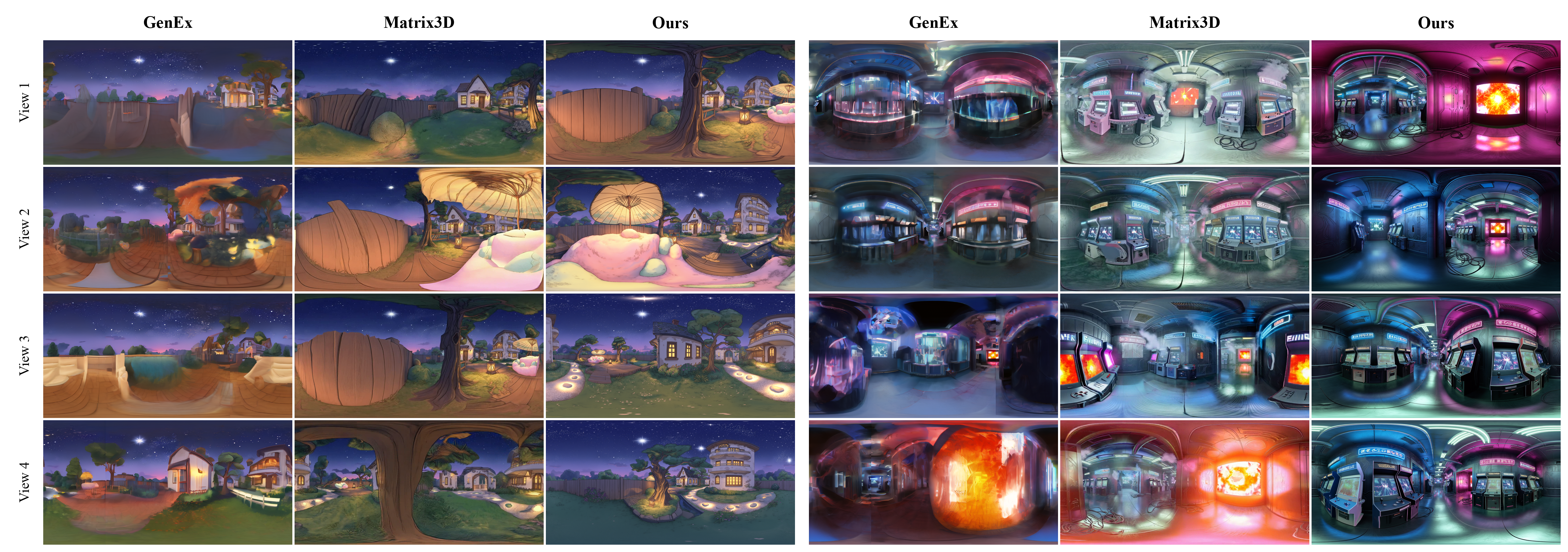}
    \caption{\textbf{Qualitative comparison on in-the-wild panoramas.}
    A single video trajectory cannot reach all four directional target viewpoints, so GenEx~\citep{lu2025genex} and Matrix-3D~\citep{yang2025matrix} each generate four independent clips. Their results exhibit ERP distortions, content collapse on diverse wild inputs, and abnormal overexposure, whereas WorldExpand preserves coherent structure and geometric cues across different visual styles. Additional comparisons and analysis are provided in Supplementary \Cref{sec:matrix3d_system_comparison,app:qual,app:realsee}.
    \label{fig:pano_result}}
\end{figure*}

\noindent\textbf{Qualitative Comparison and Reconstruction Analysis.}
\Cref{fig:pano_result} evaluates in-the-wild inputs with diverse scene styles. Because the video baselines require a separate trajectory for each direction, their four clips are generated independently and can suffer ERP distortion, severe appearance drift, content collapse, or unstable exposure. WorldExpand instead generates the four target panoramas jointly and remains robust across these styles while preserving clearer structure and geometric cues. The downstream point-cloud comparison in \Cref{fig:pcd_compare} further shows that these consistent observations produce stable, well-defined 3D structures, whereas Matrix-3D exhibits local collapse and distorted geometry. Controlled comparisons on UE synthetic and RealSee3D, together with further system-level analysis, are provided in Supplementary \Cref{sec:matrix3d_system_comparison,app:qual,app:realsee}.

\begin{figure*}[t]
    \centering
    \includegraphics[width=\linewidth]{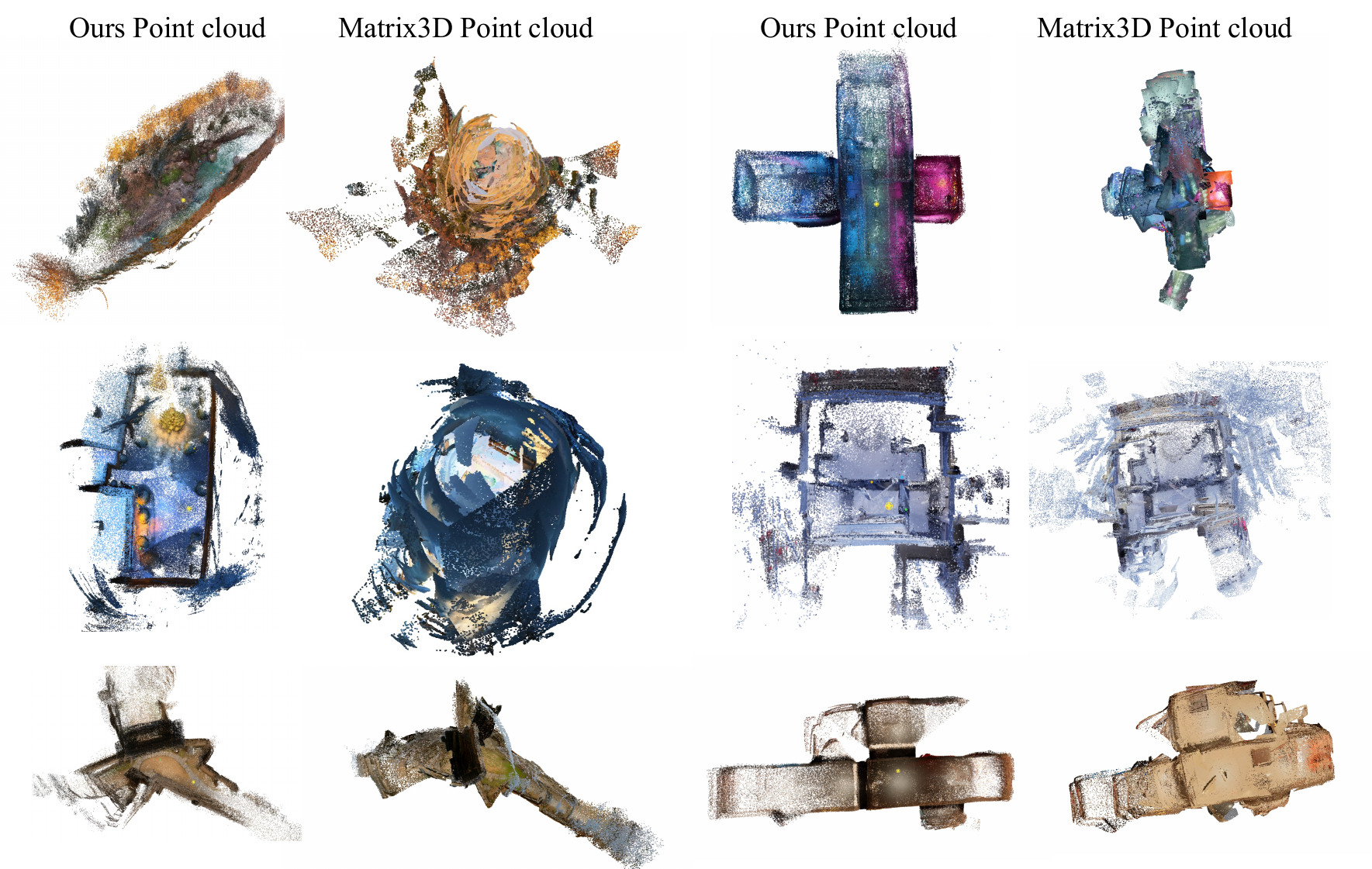}
    \caption{\textbf{Additional in-the-wild 3D reconstruction comparisons.} Across diverse panoramic inputs, WorldExpand produces stable, well-defined point-cloud structures. In contrast, Matrix-3D exhibits local point-cloud collapse and distorted geometric details, reflecting inconsistencies in its generated observations.}
    \label{fig:pcd_compare}
\end{figure*}

\subsection{Ablation Study}

We ablate the key components of WorldExpand to assess their contributions to view coverage, cross-view consistency, generation quality, and efficiency. Supplementary \Cref{app:ablation,sec:local_layer_tradeoff} give further results for joint generation and local attention.

\noindent\textbf{Diagnostic-Guided Global--Local Attention Schedule.}
Prior panoramic-video methods allocate their computational budget to densely sampled, highly redundant frames, leaving limited memory for high spatial resolution. Our sparse panoramic keyframes instead concentrate this budget on a few high-resolution observations, yielding up to an $8.6\times$ speed-up for four-direction generation over the panoramic-video baseline; detailed system-level timings are provided in Supplementary \Cref{sec:matrix3d_system_comparison}. However, joint 2K synthesis remains expensive if every transformer layer performs full cross-view attention. We therefore ablate how the number of diagnostic-selected local layers affects training and inference speed, memory, and generation quality. As shown in \Cref{fig:local_attn_tradeoff}, localizing more layers steadily reduces cost with only a modest quality trade-off; using 30 of 60 layers as local provides a balanced operating point. Supplementary \Cref{sec:local_layer_tradeoff} reports the measurements for each layer count.

\begin{figure}[t]
\centering
\includegraphics[height=4.05cm]{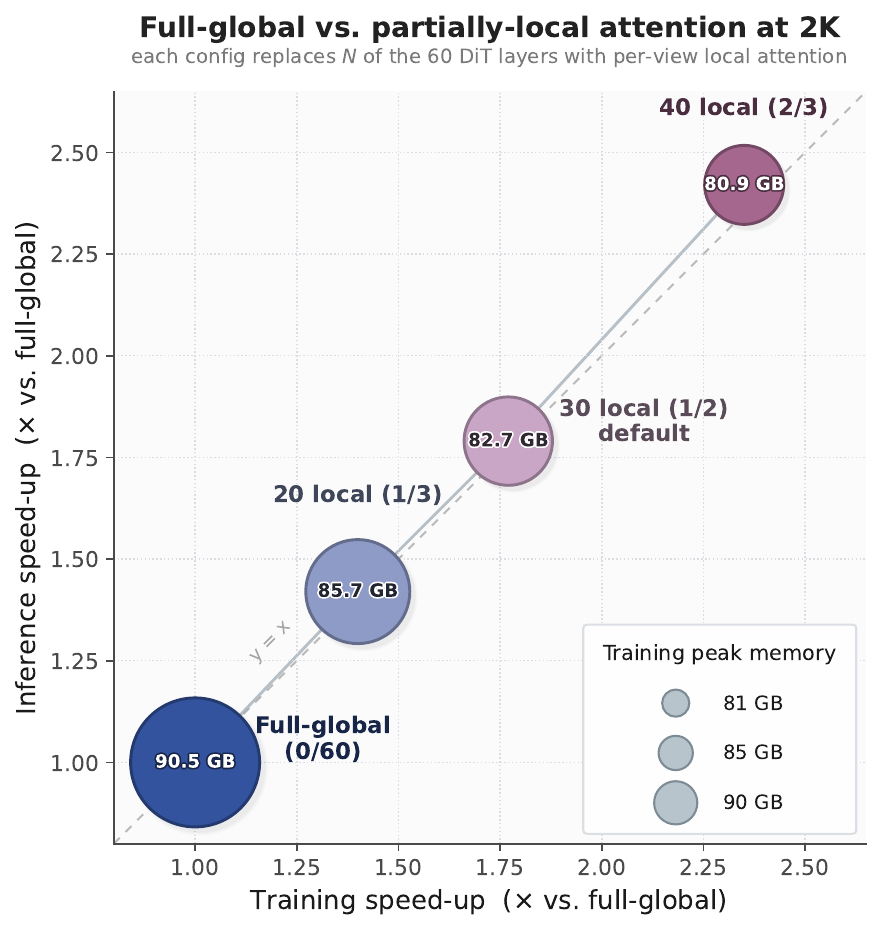}\hspace{0.02\linewidth}\includegraphics[height=4.05cm]{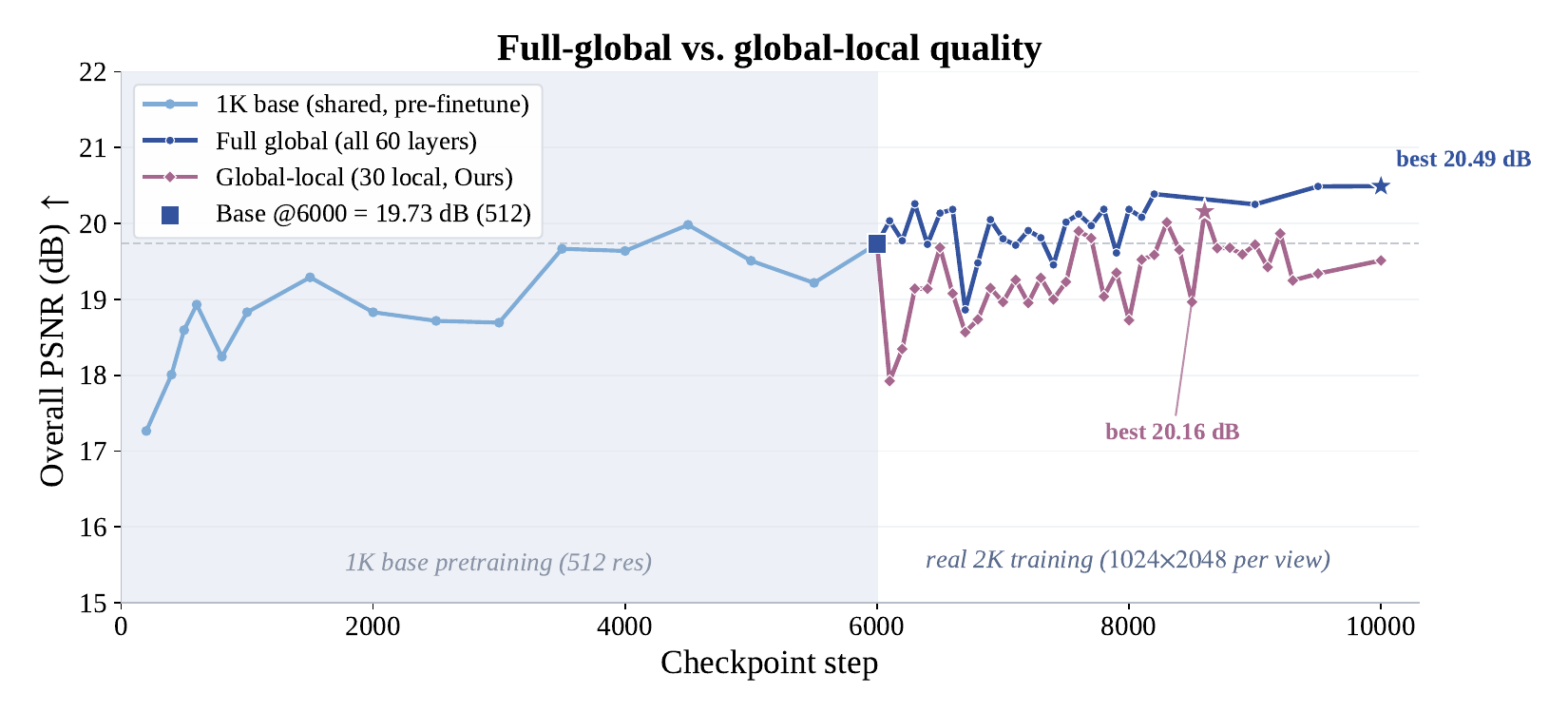}
\caption{\textbf{Global--local attention is an efficiency--quality trade-off.} \emph{Left}: training vs.\ inference speed-up over the full-global baseline at 2K; bubble size is the training peak memory (localizing more layers is faster and lighter). \textbf{Our default $30$-local schedule reaches nearly $2\times$ speed-up in both training and inference ($1.77\times$/$1.79\times$) while cutting peak memory.} \emph{Right}: PSNR vs.\ checkpoint step for full-global vs.\ our default $30$-local schedule---the two track within ${\sim}0.33$~dB. Full analysis in Supplementary \Cref{sec:local_layer_tradeoff}. \label{fig:local_attn_tradeoff}}
\vspace{-6pt}
\end{figure}

\noindent\textbf{Joint Four-View Generation.}
In \emph{w/o four-view}, we train a single-view variant that accepts only one target warp at a time and independently denoises the four target panoramas without cross-view attention. Qualitatively, joint generation produces more consistent appearance and structure across overlapping regions, as shown in \Cref{fig:joint_four_view_main}; additional examples are provided in Supplementary \Cref{fig:abl}. Quantitatively, it raises the four-view average PSNR from $18.27$ to $20.16$~dB ($+1.89$~dB) and SSIM from $0.560$ to $0.658$ ($+0.098$), while reducing LPIPS from $0.284$ to $0.256$ ($-0.028$; Supplementary \Cref{app:ablation,tab:ablation}). Because the ground-truth panoramas are mutually consistent across viewpoints, these gains provide indirect evidence that the generated set more faithfully preserves shared scene content.

\begin{figure}[t]
    \centering
    \includegraphics[width=\linewidth]{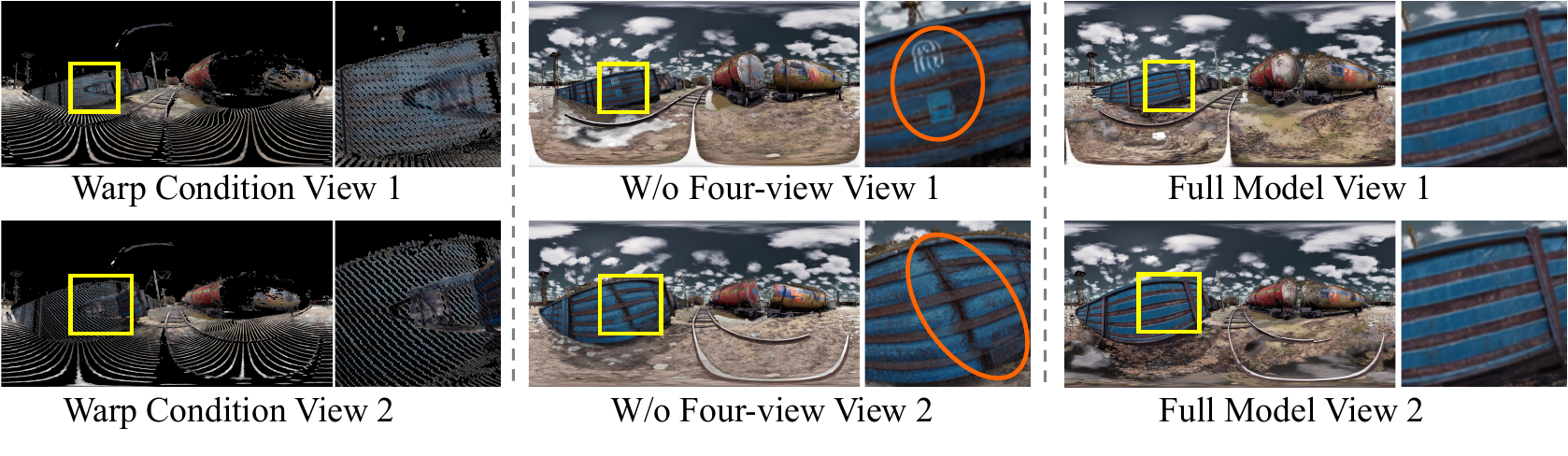}
    \caption{\textbf{Joint four-view generation improves consistency.} Independently generated views disagree in an overlapping structure (orange ellipses), whereas joint generation preserves consistent content across the two views.}
    \label{fig:joint_four_view_main}
\end{figure}

\noindent\textbf{Geometric Conditioning Design.}
We condition on a colored point cloud rendered by Gaussian splatting. Splatting is $\sim$$5$--$6\times$ faster than plain point rendering and orders of magnitude faster than meshing (\Cref{tab:condition_main}, left), and its cleaner warp---especially where the point cloud is locally sparse---also yields better synthesis quality than plain point rendering ($+0.76$~dB PSNR; \Cref{tab:condition_main}, right). We prefer a point cloud over a reconstructed mesh, which stretches ``rubber-sheet'' triangles across depth discontinuities and cannot be augmented, and over depth-as-color, which loses foreground contrast under the large outdoor depth range.

\begin{table}[t]
\centering
\caption{\textbf{Point-cloud warp condition rendered by Gaussian splatting.} \emph{Left}: render cost for the $N{=}4$ target views at 1K/2K per-view resolution (splatting is much cheaper and point-count independent). \emph{Right}: synthesis quality of splatting vs.\ plain point rendering (same point cloud, identical protocol). Best in \textbf{bold}.}
\label{tab:condition_main}
\footnotesize
\begin{minipage}[b]{0.53\linewidth}
\centering
\setlength{\tabcolsep}{3pt}
\begin{tabular}{lcc}
\toprule
Render engine & 1K & 2K \\
\midrule
Gaussian splatting (ours) & \textbf{0.72\,s} & \textbf{0.89\,s} \\
Point cloud (1M pts) & 4.21\,s & 4.51\,s \\
Mesh & 60\,s & $\sim$73\,s \\
\bottomrule
\end{tabular}
\end{minipage}\hfill
\begin{minipage}[b]{0.45\linewidth}
\centering
\setlength{\tabcolsep}{3pt}
\begin{tabular}{lccc}
\toprule
Renderer & PSNR$\uparrow$ & SSIM$\uparrow$ & LPIPS$\downarrow$ \\
\midrule
Point render & 19.40 & 0.593 & \textbf{0.252} \\
Splatting (ours) & \textbf{20.16} & \textbf{0.658} & 0.256 \\
\bottomrule
\end{tabular}
\end{minipage}
\vspace{-6pt}
\end{table}

\par\endgroup
\section{Conclusion}

We presented WorldExpand for single-panorama 3D scene expansion through sparse, complementary high-resolution observations. Coverage-guided selection, joint panoramic diffusion, and diagnostic-guided attention address incomplete spatial coverage, cross-view inconsistency, and generation cost. Experiments demonstrate improved novel-view synthesis and reconstruction, with up to an $8.6\times$ denoising speed-up over the state-of-the-art panoramic-video method. These advances address substantial challenges that remain in coherent, detailed 3D world generation.

\paragraph{Limitations and Future Work.}
We focus on static 3D scene generation, where complete spatial coverage, fine detail, and geometric consistency remain open challenges. Extending spatial expansion to dynamic 4D environments, including dynamic-object insertion while maintaining geometric and temporal consistency, is an interesting direction for future research.

\subsection*{AI Use Statement}
We used generative AI tools to assist with language editing and manuscript formatting. All AI-assisted text and formatting changes were reviewed by the authors, who verified the technical claims and take responsibility for the final content.

\bibliography{ref}
\bibliographystyle{iclr2027_conference}

\clearpage
\appendix
\setlength{\textfloatsep}{8pt plus 2pt minus 2pt}
\setlength{\floatsep}{8pt plus 2pt minus 2pt}
\setlength{\intextsep}{8pt plus 2pt minus 2pt}
\setlength{\abovecaptionskip}{4pt}
\setlength{\belowcaptionskip}{0pt}

\section*{Appendix Overview}
This supplementary material is organized as follows:
\begin{itemize}
    \item \Cref{sec:implementation_details}: training and reconstruction implementation details.\dotfill p.~\pageref{sec:implementation_details}
    \item \Cref{sec:reconstruction_details}: perspective decomposition, known-camera depth alignment, Gaussian initialization, and optimization details.\dotfill p.~\pageref{sec:reconstruction_details}
    \item \Cref{sec:matrix3d_system_comparison}: system-level comparison with Matrix-3D.\dotfill p.~\pageref{sec:matrix3d_system_comparison}
    \item \Cref{app:dataset}: details of scene collection, roaming-trajectory generation, and intersection-centric anchor--target sampling for the synthetic dataset.\dotfill p.~\pageref{app:dataset}
    \item \Cref{app:ablation}: quantitative and qualitative ablation on the joint four-view generation design.\dotfill p.~\pageref{app:ablation}
    \item \Cref{sec:local_layer_tradeoff}: trade-off between the number of local-attention layers and the training memory/speed.\dotfill p.~\pageref{sec:local_layer_tradeoff}
    \item \Cref{app:per_view_results}: full per-view NVS results on UE synthetic and RealSee3D.\dotfill p.~\pageref{app:per_view_results}
    \item \Cref{app:qual}: qualitative comparisons on held-out UE synthetic data.\dotfill p.~\pageref{app:qual}
    \item \Cref{app:realsee}: additional qualitative novel-view synthesis comparisons on the RealSee3D dataset.\dotfill p.~\pageref{app:realsee}
\end{itemize}

\section{Implementation Details}
\label{sec:implementation_details}
\noindent\textbf{Setup.}
We build our novel-view panorama generator on a pretrained image diffusion model~\citep{wu2025qwen} and train it in two stages for joint multi-view panoramic synthesis, with each training and inference step jointly producing four novel-view panoramas. We first train at 1K resolution ($512 \times 1024$ per view) with full global attention for $6{,}000$ steps, which teaches the model to follow the point-cloud warp condition and perform novel-view synthesis; we then fine-tune for $2{,}000$ steps at 2K resolution ($1024 \times 2048$ per view) with the global--local attention schedule that switches $30$ of the $60$ DiT layers to per-view local attention. Thus, all reported main-model trajectories switch from 1K to 2K at step $6{,}000$. Both stages use a batch of eight training examples per iteration. For the feed-forward reconstruction stage, we adopt WorldMirror2~\citep{hy2026hy} as the perspective backbone. Each input panorama is sliced into 21 perspective views that uniformly cover the $360^\circ$ field of view, which are fed into the feed-forward model to obtain globally consistent depth predictions. After known-camera calibration and RANSAC-based scale alignment, the resulting point cloud is used to initialize a Gaussian Splatting~\citep{kerbl20233d} representation, and the final scene is optimized using a gsplat-based~\citep{ye2025gsplat} implementation.

\subsection{Reconstruction Details}
\label{sec:reconstruction_details}
We follow the WorldMirror 2.0 and 3DGS composition pipeline of HY-World 2.0~\citep{hy2026hy} to convert the generated posed panoramas into an explicit 3D representation. The details below are included for completeness; reconstruction is not a contribution of our method.

\noindent\textbf{Perspective Decomposition and Depth Prediction.}
Let the full panorama set be $\mathcal{I}=\{I_0\}\cup\{\hat{I}_i\}_{i=1}^{N}$ with known camera poses $\{\mathbf{T}_i\}_{i=0}^{N}$. For each panorama, we sample $K{=}21$ tangent cameras that uniformly cover the sphere and render dense perspective crops. We jointly feed all crops into WorldMirror 2.0~\citep{liu2025worldmirror}, which performs cross-view reasoning and predicts per-crop depths $\hat{D}_{i,k}$, where $k\in\{1,\ldots,K\}$. We discard its predicted cameras and retain the globally coherent depth predictions because the panorama poses are already known.

\noindent\textbf{Known-Camera Depth Alignment and Gaussian Initialization.}
Each tangent view has a known camera pose $\mathbf{T}_{i,k}$, obtained by composing panorama pose $\mathbf{T}_i$ with the $k$-th tangent-camera rotation. Let $\Omega_{i,k}^{\mathrm{align}}$ denote pixels with valid correspondences between predicted depth $\hat{D}_{i,k}$ and anchor depth $D_{i,k}^{\mathrm{anchor}}$. For the source panorama ($i{=}0$), the anchor is obtained by projecting its depth $D_0$ into the tangent view; for generated panoramas ($i{\geq}1$), it is obtained by warping the source geometry into the corresponding target tangent view. We pool all valid correspondences and estimate one scene-level scale shared by all crops:
\begin{equation}
    s^{\star}=\operatorname*{RANSAC}_{\substack{i\in\{0,\ldots,N\},\,k\in\{1,\ldots,K\}\\\mathbf{x}\in\Omega_{i,k}^{\mathrm{align}}}}
    \frac{D_{i,k}^{\mathrm{anchor}}(\mathbf{x})}{\hat{D}_{i,k}(\mathbf{x})}.
\end{equation}
Using a shared scale preserves the relative geometry predicted across panoramas and avoids independently scaled local reconstructions. Let $\Omega_{i,k}^{\mathrm{recon}}$ contain all valid, non-sky depth predictions retained for reconstruction, including newly revealed regions. We apply the scale to all predicted depths and back-project them using the known cameras:
\begin{equation}
\tilde{D}_{i,k}=s^{\star}\hat{D}_{i,k},\quad
\mathcal{P}_{i,k}=\left\{\mathbf{R}_{i,k}^{\top}\!\left(\tilde{D}_{i,k}(\mathbf{x})\,\mathbf{K}^{-1}\bar{\mathbf{x}}-\mathbf{t}_{i,k}\right)\mid\mathbf{x}\in\Omega_{i,k}^{\mathrm{recon}}\right\},
\end{equation}
where $\bar{\mathbf{x}}$ is the homogeneous pixel coordinate, $\mathbf{K}$ is the tangent-camera intrinsic matrix, and $(\mathbf{R}_{i,k},\mathbf{t}_{i,k})$ are the rotation and translation components of $\mathbf{T}_{i,k}$. Merging all local point clouds gives $\mathcal{P}_{\mathrm{init}}=\bigcup_{i,k}\mathcal{P}_{i,k}$, which initializes the Gaussian representation and includes geometry inferred in newly exposed regions.

\noindent\textbf{Evaluation Depth Protocol.}
For a controlled comparison in \Cref{tab:main_results}, every method uses the same source-view depth wherever geometry is required for view planning, point-cloud warp construction, or final scale alignment. We use rendered ground-truth source depth on both UE synthetic and the RealSee3D synthetic subset. GenEx, Matrix-3D, and WorldExpand are therefore evaluated with identical source geometry and identical selected target poses; no target-view depth is provided to any method, and target RGB-D is used only to compute evaluation metrics. For the in-the-wild qualitative examples, where ground-truth depth is unavailable, we instead predict the source depth using DepthMaster~\citep{wang2026depthmaster}. This separates the controlled benchmark protocol from the practical deployment setting with estimated geometry.

\noindent\textbf{Pano-Aware Gaussian Optimization.}
Starting from the initialized point cloud $\mathcal{P}_{\mathrm{init}}$, we follow the Gaussian optimization design of HY-World 2.0~\citep{hy2026hy} and optimize a 3D Gaussian Splatting representation $\mathcal{G}$ under sparse panoramic supervision. For each tangent training view, the differentiable rasterizer outputs rendered color $\bar{I}_{i,k}$, rendered inverse depth $\bar{D}_{i,k}^{-1}$, and rendered normals $\bar{\mathbf{N}}_{i,k}$. The optimization combines color, geometry, and reliability terms (view subscripts omitted for brevity):
\begin{align}
\mathcal{L}={}&\lambda_{\mathrm{rgb}}\mathcal{L}_{\mathrm{rgb}}
+\lambda_{\mathrm{lpips}}\mathcal{L}_{\mathrm{lpips}}
+\lambda_{\mathrm{disp}}\left\lVert\bar{D}^{-1}-\tilde{D}^{-1}\right\rVert_1 \nonumber\\
&+\lambda_{\mathrm{n}}\left(1-\left\langle\bar{\mathbf{N}},\mathbf{N}^{\mathrm{pseudo}}\right\rangle\right)
+\lambda_{\mathrm{mask}}\mathcal{L}_{\mathrm{mask}}.
\end{align}
We use inverse-depth supervision because panoramic scenes span extreme depth ranges, and derive $\mathbf{N}^{\mathrm{pseudo}}$ from the aligned depth $\tilde{D}$ by finite differences. The mask loss is $\mathcal{L}_{\mathrm{mask}}=(\operatorname{mean}(\bar{M}))^2$, where $\bar{M}$ is the rendered Gaussian mask, and suppresses diffuse opacity. We additionally place virtual cameras between selected panorama pairs and supervise their rendered inverse depth using pseudo-depth warped from nearby training views, regularizing geometry between the sparse observations.

We optimize the Gaussian representation for $T{=}1{,}500$ steps. The color term is $\mathcal{L}_{\mathrm{rgb}}=0.8\lVert\bar{I}-I\rVert_1+0.2(1-\operatorname{SSIM}(\bar{I},I))$ throughout optimization. For the remaining terms, each weight is linearly interpolated as $\lambda(t)=(1-t/T)\lambda^{\mathrm{start}}+(t/T)\lambda^{\mathrm{end}}$: $\lambda_{\mathrm{lpips}}$ decreases from $0.2$ to $0.1$, $\lambda_{\mathrm{disp}}$ decreases from $0.2$ to $0.01$, and $\lambda_{\mathrm{n}}$ increases from $0.1$ to $0.5$. The inverse-depth residual is additionally normalized by the scene scale. We keep $\lambda_{\mathrm{mask}}=0.003$ and activate the mask term only from step $300$ onward. These schedules were selected empirically to balance appearance fitting and geometric refinement rather than through exhaustive tuning. In practice, the reconstruction is insensitive to modest perturbations of these values, which produce no material change in the resulting geometry or rendering quality.

\section{System-Level Comparison with Matrix-3D}
\label{sec:matrix3d_system_comparison}
We compare WorldExpand with Matrix-3D~\citep{yang2025matrix} in \Cref{tab:matrix3d_system_comparison}. We use the same setup for all runtime measurements. WorldExpand uses a 20B-parameter backbone and generates panoramas at $1024{\times}2048$, whereas Matrix-3D uses a 14B-parameter backbone and operates at $480{\times}960$ or $720{\times}1440$. Despite using the larger model and higher output resolution, WorldExpand completes four-direction denoising approximately $2.3\times$ faster than Matrix-3D at 480p and $8.6\times$ faster at 720p. This advantage comes from its reconstruction-oriented design: it spends computation on sparse complementary observations, generates them jointly, and directly fuses their posed panoramas. Matrix-3D instead generates a dense 81-frame video for each direction. Most adjacent frames provide little new scene coverage, yet substantially increase generation and downstream processing cost; exploring four directions further requires four independent runs without cross-trajectory communication. Its generation stage is therefore not co-designed to provide a compact multi-direction observation set for reconstruction, and its official reconstruction interface processes one video trajectory at a time.

\begin{table*}[t]
\centering
\caption{\textbf{System-level comparison with Matrix-3D.} Despite using a larger model and higher resolution, WorldExpand is approximately $2.3\times$/$8.6\times$ faster than Matrix-3D at 480p/720p for four-direction generation. We multiply the inference time of one Matrix-3D clip by four to account for the four independent directions. The denoising row is stage-matched, whereas WorldExpand end-to-end time was not separately profiled. Render costs follow the controlled comparison in \Cref{tab:condition_main}.}
\label{tab:matrix3d_system_comparison}
\small
\setlength{\tabcolsep}{3pt}
\begin{tabular}{p{0.20\textwidth} p{0.37\textwidth} p{0.37\textwidth}}
\toprule
\textbf{Property} & \textbf{WorldExpand} & \textbf{Matrix-3D} \\
\midrule
\textbf{Generation backbone}
& 20B
& 14B \\

\textbf{Resolution}
& $1024{\times}2048$
& $480{\times}960$ or $720{\times}1440$ \\

\textbf{Representation}
& 4 joint panoramas
& 81-frame video per direction \\

\textbf{Four directions}
& 1 run; 4 panoramas
& 4 runs; 324 frames \\

\textbf{Denoising (4 directions)}
& ${\sim}45$\,min $26$\,s (1 joint run)
& $1$\,h $45$\,min $42$\,s (4 clips, 480p) / $6$\,h $29$\,min $53$\,s (4 clips, 720p) \\

\textbf{Condition}
& Gaussian-splat point cloud
& MoGe-derived mesh \\

\textbf{Render cost}
& GS: $0.72$\,s / $0.89$\,s (1K / 2K)
& Mesh ref.: $60$\,s / ${\sim}73$\,s (1K / 2K) \\

\textbf{Consistency}
& Joint across views
& Independent across directions \\

\textbf{Reconstruction}
& Arbitrary posed panoramas
& Single trajectory; PanoLRM/Pano-GS \\
\bottomrule
\end{tabular}
\end{table*}

\begin{figure}[p]
    \centering
    \includegraphics[width=\linewidth,height=0.88\textheight,keepaspectratio]{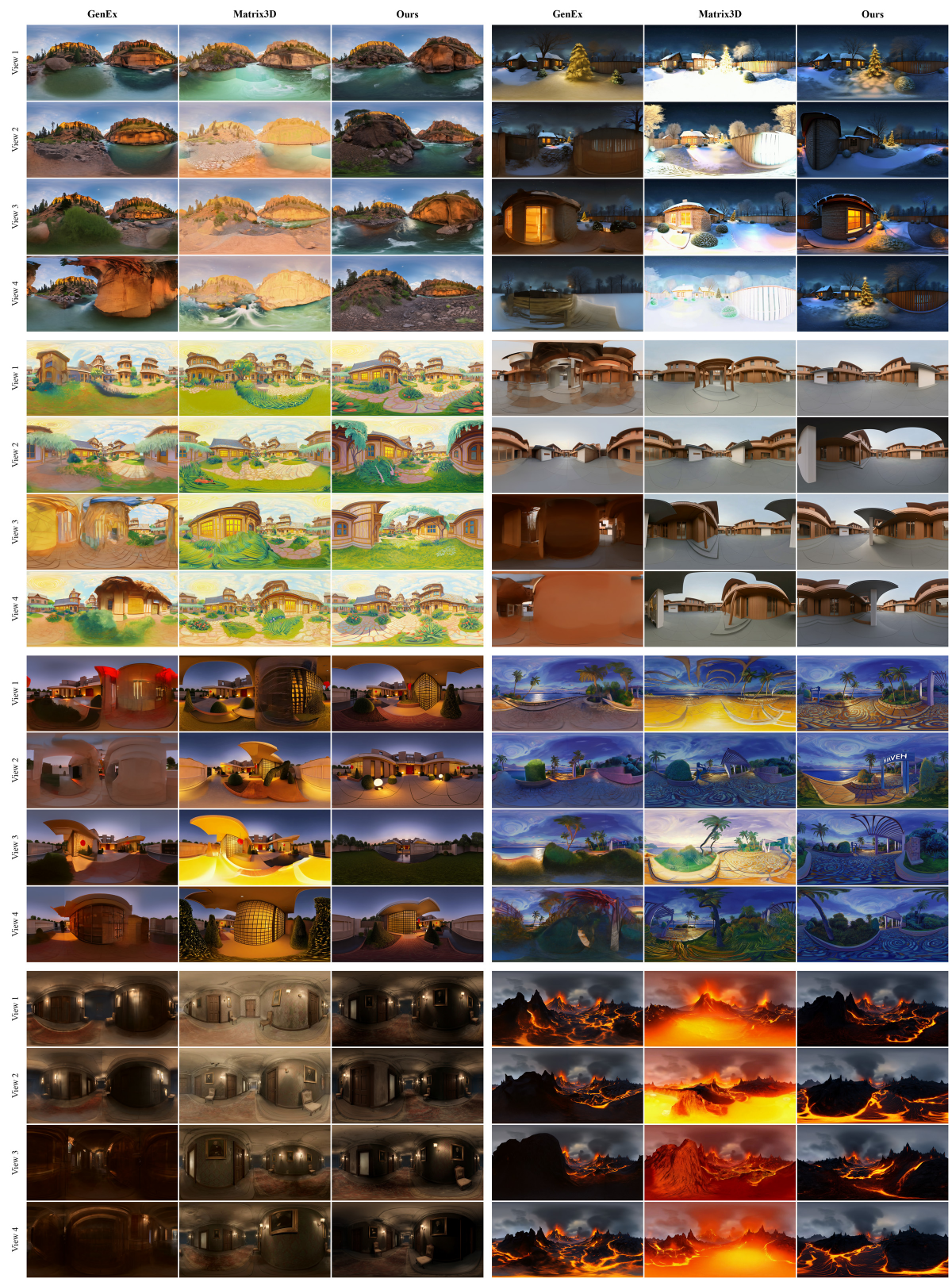}
    \caption{\textbf{Additional in-the-wild comparisons.} The four groups are arranged on one page; method headers are shared at the top. WorldExpand remains robust across diverse styles, while independent video generation is more susceptible to distortion, exposure shifts, and structural degradation.}
    \label{fig:matrix3d_wild_2_3}
\end{figure}

\noindent\textbf{Additional In-the-Wild Comparisons.}
\Cref{fig:matrix3d_wild_2_3} extends the main-paper comparison to more architectural, game-like, stylized, indoor, and outdoor panoramas. Across these diverse inputs, independently generated video trajectories frequently introduce ERP distortion, unstable exposure, or content collapse, whereas WorldExpand retains clearer structure and geometric cues across the four target views.

\begin{figure}[p]
    \centering
    \includegraphics[width=\linewidth]{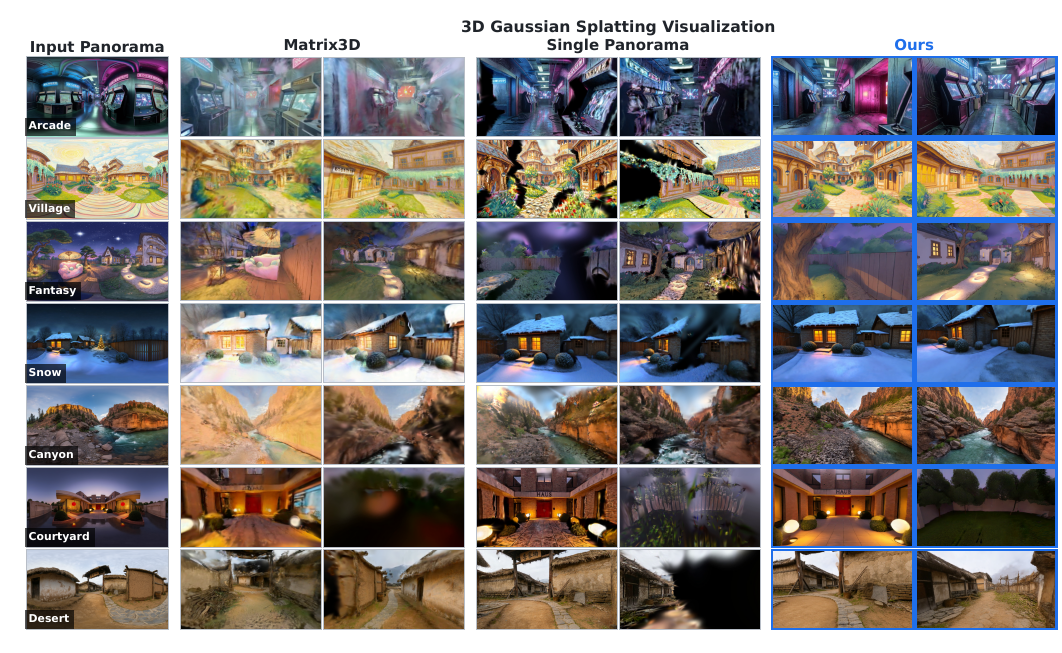}
    \caption{\textbf{Additional in-the-wild 3D Gaussian Splatting comparisons.} Each row shows the input panorama followed by reconstructions obtained with Matrix-3D, the single panorama alone, and WorldExpand. Across diverse scenes, WorldExpand provides broader scene coverage and more stable geometry, while single-panorama reconstructions remain incomplete and Matrix-3D exhibits distorted or inconsistent structure.}
    \label{fig:matrix3d_single_ours_recon}
\end{figure}

\section{Dataset Construction Details}
\label{app:dataset}
Training a pose-controllable multi-pano generation model requires scenes with multiple walking directions, moderate anchor--target disparity, accurate camera poses, and dense depth for warp-based conditioning. Real-world panoramic data cannot reliably provide these properties due to motion artifacts, missing metric depth, and limited trajectory diversity; we therefore construct a large-scale synthetic dataset in Unreal Engine. Each training tuple is
\begin{equation}
\mathcal{S}=\left(I_0,D_0,\mathbf{T}_0,\{I_i,\Delta\mathbf{T}_{0\rightarrow i}\}_{i=1}^{N}\right),
\end{equation}
where $I_0$, $D_0$, and $\mathbf{T}_0$ are the anchor panorama, depth, and camera pose, while $I_i$ and $\Delta\mathbf{T}_{0\rightarrow i}$ are the $i$-th target panorama and its relative pose. We next detail the scene collection, trajectory generation, and anchor--target sampling used to construct these tuples.

\noindent\textbf{Scene and Roaming Trajectory Collection.}
We curate Unreal Engine scenes with complex meshes and diverse layouts, and build a trajectory generation pipeline. For each scene, we construct navigable regions via NavMesh-based collision avoidance, generate camera paths with Unreal Engine's random roaming, and filter out invalid cases such as collisions, mesh penetration, short movements, and low-parallax trajectories. We then render equirectangular RGB and depth from a panoramic camera along each valid trajectory, mark sky regions as infinite depth, and record synchronized RGB, depth, and pose annotations. The final dataset contains $76{,}721$ individual trajectories, corresponding to approximately $19.2$K four-trajectory groups, covering diverse camera displacements and reliable geometric correspondences.
We reserve $30$ scenes for validation, with the remaining scenes used for training, ensuring no scene-level overlap between splits.

\noindent\textbf{Intersection-Centric NVS Pair Collection.}
Exhaustively enumerating anchor--target combinations would produce many redundant pairs: very close views have little parallax, while very distant views encourage hallucination. We instead use an intersection-aware sampling strategy. For each scene, we project all camera centers onto the ground plane and detect intersections between trajectory segments. Nearby intersections are merged, and the frame closest to each merged point is used as an anchor panorama. These intersections naturally connect multiple walking directions and provide a reconstruction-oriented center for scene expansion. Around each anchor, we sample four target views from the incoming and outgoing trajectory directions, subject to a valid baseline range and a minimum temporal gap. For each anchor--target pair, we unproject the anchor RGB-D panorama into a colored point cloud, transform it by the relative pose, and render the point cloud at the target view to obtain a warped image and a disocclusion mask that explicitly marks regions requiring synthesis. This intersection-centered design yields approximately $120$K effective NVS training tuples.

\begin{figure*}[htbp]
    \centering
    \includegraphics[width=1.0\linewidth]{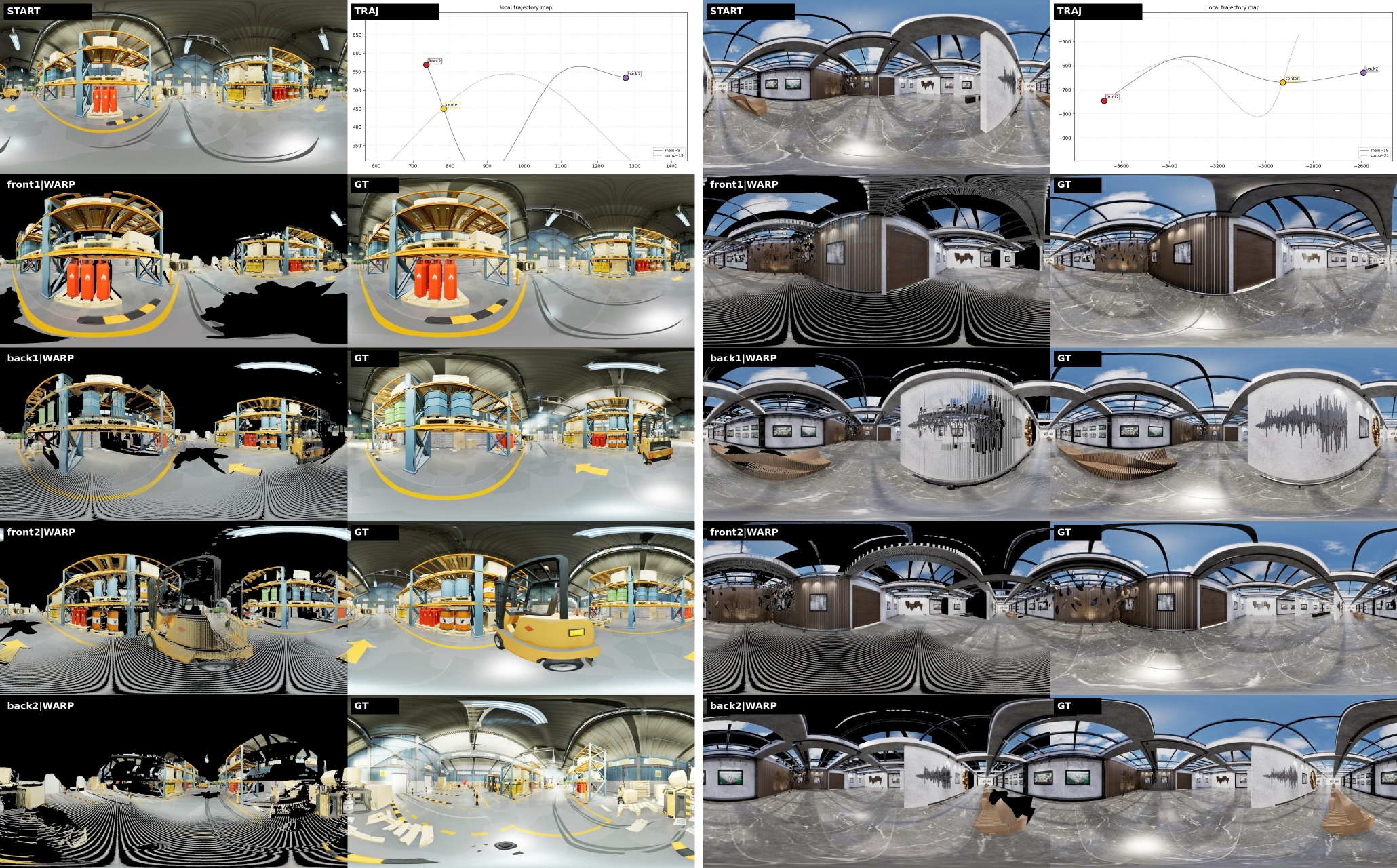}
    \caption{\textbf{Dataset examples.} Two training samples from our synthetic Unreal Engine dataset, an outdoor \emph{Colonial City} scene (left) and an indoor \emph{gallery} scene (right). For each example we show the anchor panorama (\textsc{start}) and its local roaming trajectory (\textsc{traj}), followed by the four target views: the point-cloud warp condition (\textsc{warp}) and the corresponding ground-truth panorama (\textsc{gt}). The warps expose disocclusions (black regions) that the model must synthesize while staying consistent with the anchor. \label{fig:dataset_examples}}
\end{figure*}

\FloatBarrier
\section{Ablation on Joint Four-View Generation}
\label{app:ablation}
\begin{figure}[!ht]
    \centering
    \begin{minipage}{1\linewidth}
    \centering
    \parbox[b]{0.326\linewidth}{\centering\footnotesize\textbf{Warped geometry condition}\\(shared input)}%
    \parbox[b]{0.326\linewidth}{\centering\footnotesize\textbf{W/o Four-View}\\independent per-view generation $\rightarrow$ \emph{inconsistent} across views}%
    \parbox[b]{0.326\linewidth}{\centering\footnotesize\textbf{Full Model}\\joint four-view generation $\rightarrow$ \emph{consistent} across views}\\[2pt]
    \includegraphics[width=\linewidth]{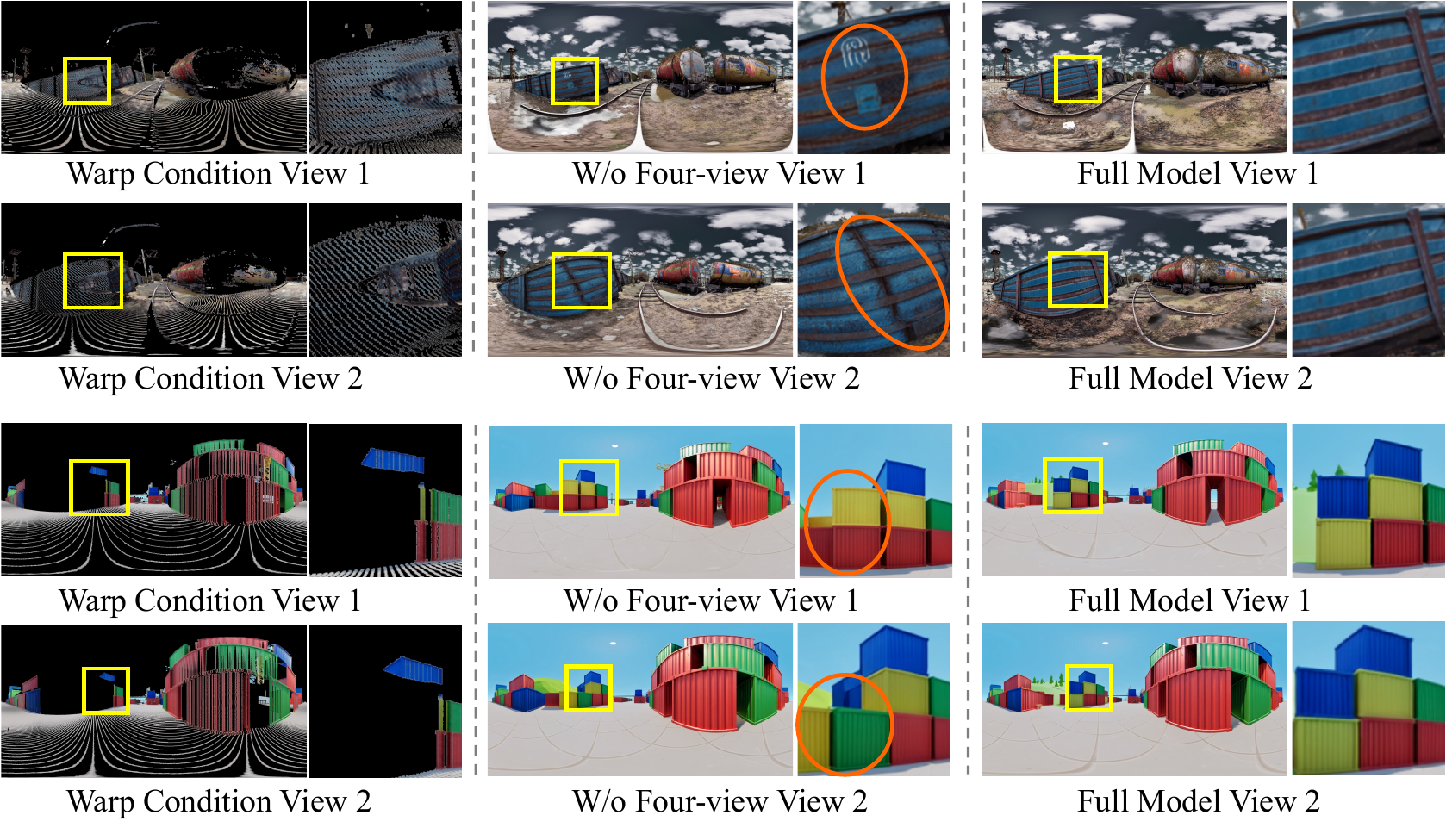}
    \end{minipage}
    \caption{\textbf{Joint four-view generation improves consistency.} Independently generated views disagree in overlapping regions (orange ellipses), whereas joint generation produces coherent content.}
    \label{fig:abl}
\end{figure}

\Cref{tab:ablation} and \Cref{fig:abl} provide the quantitative and qualitative ablations of joint four-view generation, respectively. When the target views are generated independently (single-view generation) instead of jointly, all per-view metrics degrade and the model produces inconsistent content across views over multiple inference runs. Together, these results confirm that joint generation is essential for both per-view fidelity and cross-view consistency.

\begin{table}[h!]
\centering
\caption{\textbf{Ablation on joint four-view generation.} \emph{w/o four-view} trains a single-view variant whose conditioning inputs support only one target warp at a time, so the four target panoramas are denoised independently at inference. \Cref{fig:local_attn_tradeoff} reports the efficiency and quality of the global--local attention schedule. Best results per metric are in \textbf{bold}.}
\label{tab:ablation}
\setlength{\tabcolsep}{4pt}
\begin{tabular}{lcccccc}
\toprule
& \multicolumn{3}{c}{w/o four-view} & \multicolumn{3}{c}{Full} \\
\cmidrule(lr){2-4}\cmidrule(lr){5-7}
View & PSNR $\uparrow$ & SSIM $\uparrow$ & LPIPS $\downarrow$
     & PSNR $\uparrow$ & SSIM $\uparrow$ & LPIPS $\downarrow$ \\
\midrule
V1   & 19.16 & 0.589 & 0.251 & \textbf{20.61} & \textbf{0.674} & \textbf{0.243} \\
V2   & 18.20 & 0.558 & 0.290 & \textbf{20.05} & \textbf{0.663} & \textbf{0.256} \\
V3   & 18.51 & 0.570 & 0.281 & \textbf{19.91} & \textbf{0.647} & \textbf{0.262} \\
V4   & 17.23 & 0.523 & 0.315 & \textbf{20.06} & \textbf{0.649} & \textbf{0.261} \\
Avg. & 18.27 & 0.560 & 0.284 & \textbf{20.16} & \textbf{0.658} & \textbf{0.256} \\
\bottomrule
\end{tabular}
\end{table}

\FloatBarrier
\section{Trade-off Between Number and Cost of Attention Layers}
\label{sec:local_layer_tradeoff}
The global--local schedule has two design choices: the token scope visible to each local layer and the number of layers made local. We first select the most efficient local token scope, then fix that design and study the layer-count trade-off.

\noindent\textbf{Selecting the Local Token Scope.}
We compare three per-view token scopes while leaving the global layers unchanged:
\begin{equation}
\begin{aligned}
\text{v1:}\quad \mathbf{H}^{\mathrm{local,v1}}_i
    &= [\mathbf{P}_{txt};\mathbf{Z}_i],\\
\text{v2:}\quad \mathbf{H}^{\mathrm{local,v2}}_i
    &= [\mathbf{P}_{txt};\mathbf{Z}_i;\mathbf{W}_i],\\
\text{v3:}\quad \mathbf{H}^{\mathrm{local,v3}}_i
    &= [\mathbf{P}_{txt};\mathbf{S}_0;\mathbf{Z}_i;\mathbf{W}_i].
\end{aligned}
\label{eq:local_attention_variants}
\end{equation}
Here $\mathbf{P}_{txt}$ denotes text tokens, $\mathbf{S}_0$ the source-panorama tokens, $\mathbf{Z}_i$ the noisy tokens of target view $i$, and $\mathbf{W}_i$ its geometric-warp tokens. The three scopes differ in how much conditioning each local layer still sees: v1 lets a local layer refine only its own target view, v2 additionally keeps that view's geometric warp inside the local layer, and v3 further keeps the shared source panorama. Every stream that is excluded locally is still available in the global layers, so the variants change \emph{where} conditioning is attended to, not whether it is used. Since each added stream lengthens the per-view local sequence, cost grows from v1 to v3, as measured at 20, 30, and 40 local layers in \Cref{tab:local_attention_variants}.

\begin{table*}[t]
\centering
\caption{\textbf{Local-attention token scopes during 2K training}. We report time per training step. OOM indicates that the configuration exceeded the available memory.}
\label{tab:local_attention_variants}
\resizebox{0.70\textwidth}{!}{%
\begin{tabular}{c l c c c c}
\toprule
Variant & Local tokens $\mathbf{H}^{\mathrm{local}}_i$ & \#Local & Peak memory & Train step & Speed-up \\
\midrule
\multirow{3}{*}{v1} & \multirow{3}{*}{$[\mathbf{P}_{txt};\mathbf{Z}_i]$}
& 20 & $85.7$\,GB & $213$\,s & $1.40\times$ \\
& & 30 & $82.7$\,GB & $169$\,s & $1.77\times$ \\
& & 40 & $80.9$\,GB & $127$\,s & $2.35\times$ \\
\midrule
\multirow{3}{*}{v2} & \multirow{3}{*}{$[\mathbf{P}_{txt};\mathbf{Z}_i;\mathbf{W}_i]$}
& 20 & $89.7$\,GB & $229$\,s & $1.31\times$ \\
& & 30 & $88.4$\,GB & $196$\,s & $1.53\times$ \\
& & 40 & $87.8$\,GB & $162$\,s & $1.85\times$ \\
\midrule
\multirow{3}{*}{v3} & \multirow{3}{*}{$[\mathbf{P}_{txt};\mathbf{S}_0;\mathbf{Z}_i;\mathbf{W}_i]$}
& 20 & OOM & -- & -- \\
& & 30 & OOM & -- & -- \\
& & 40 & OOM & -- & -- \\
\bottomrule
\end{tabular}
}
\end{table*}

Across every tested layer count, v1 gives the lowest memory and training time. Variant v2 retains more conditioning tokens locally but consequently provides much smaller savings, while v3 exceeds the available memory at all three layer counts. We therefore adopt v1: selected global layers retain source/warp-mediated cross-view reasoning, while local layers focus on per-view target refinement at the lowest computational cost.

\noindent\textbf{Varying the Number of Local Layers.}
All subsequent layer-count experiments use the v1 token scope. Following the per-layer diagnostic, we localize the \emph{least-sensitive} $N$ layers (smallest $\Delta$PSNR) rather than simply choosing the last $N$ layers. We profile training at 2K resolution ($1024{\times}2048$ per view) using DeepSpeed ZeRO-2, bf16, gradient checkpointing, LoRA (rank 64), text/VAE offloading, and a local batch size of one. We report rank-0 peak memory and training-step time (forward\,+\,backward); the full-global baseline costs $90.5$\,GB and $299$\,s per step.

\begin{table}[htbp]
\centering
\caption{\textbf{Local-attention layer-count trade-off under real 2K training}. Our backbone (Qwen-Image-Edit-2509) is a 60-layer DiT, so \#Local layers is out of 60. Baseline (all-global): $90.5$\,GB / $299$\,s per step. $\Delta$Mem is memory saved versus the baseline; speedup is relative to the baseline step time.\label{tab:local_layer_tradeoff}}
\begin{tabular}{lcccc}
\toprule
\#Local layers & Peak mem (GB) & $\Delta$Mem (GB) & Step time (s) & Speedup \\
\midrule
0 (all global) & 90.5 & --   & 299 & 1.00$\times$ \\
\midrule
20 & 85.7 & $-4.8$ & 213 & 1.40$\times$ \\
30 & 82.7 & $-7.8$ & 169 & 1.77$\times$ \\
40 & \textbf{80.9} & $\mathbf{-9.6}$ & \textbf{127} & \textbf{2.35$\times$} \\
\bottomrule
\end{tabular}
\end{table}

\noindent\textbf{Findings.} Because each local layer keeps only the per-view noisy tokens, adding more local layers steadily saves both memory and time: at 40 local layers the model reaches $80.9$\,GB with a $2.35\times$ speedup, while 20/30 local layers offer more conservative savings. The accuracy cost of localizing more layers is modest (\Cref{tab:local_layer_cost}): even at 40 local layers, the most sensitive selected layer only drops $-0.71$\,dB in the training-free diagnostic, and after finetuning the model re-adapts to local attention, so the practical degradation is smaller. \textbf{We therefore localize 30 of the 60 DiT layers by default---an empirical choice that balances efficiency and cross-view consistency---while 40 layers push efficiency further and 20 layers give a more conservative accuracy margin.}

\begin{table}[htbp]
\centering
\caption{\textbf{Accuracy cost of the selected local layers} (training-free per-layer diagnostic, overall $\Delta$PSNR; closer to $0$ is safer). Larger $N$ includes progressively more sensitive layers.\label{tab:local_layer_cost}}
\begin{tabular}{ccc}
\toprule
\#Local layers & Most sensitive selected layer ($\Delta$PSNR) & Mean $\Delta$PSNR of the set \\
\midrule
20 & $-0.19$\,dB & $-0.07$\,dB \\
30 & $-0.34$\,dB & $-0.14$\,dB \\
40 & $-0.71$\,dB & $-0.23$\,dB \\
\bottomrule
\end{tabular}
\end{table}

\noindent\textbf{Inference Speed-Up.} Local attention also shortens the per-view attention sequence at sampling time, so it accelerates inference as well (\Cref{tab:local_layer_infer_speed}): at 40 local layers the profiling latency drops from $66.8$\,s to $27.5$\,s, a $2.42\times$ speed-up, and 30 local layers (our default) already give $1.79\times$. Each reported measurement sums one conditional and one unconditional forward pass for a single denoising step. It is therefore a controlled per-step profiling proxy for comparing attention configurations, not the wall-clock runtime of a complete sampling run. Unlike training, the peak memory stays essentially flat ($\sim$$49.7$\,GB across all settings). This is expected: inference has no backward pass, so activations are freed layer-by-layer and are not the bottleneck; the peak is instead dominated by the resident model weights and the full set of image tokens ($82{,}944$ tokens for all views), which are unchanged by the attention scope. With a memory-efficient (Flash) attention kernel the attention matrix is never materialized, so even full-global attention does not inflate memory---local attention only reshapes \emph{which} tokens attend (block-diagonal per view), reducing FLOPs and latency rather than stored tensors.

\begin{table}[htbp]
\centering
\caption{\textbf{Per-step profiling speed-up from local attention at 2K} ($1024{\times}2048$ per view). Each latency sums one conditional and one unconditional forward pass over all views for a single denoising step. These controlled measurements compare attention configurations and do not represent complete end-to-end inference time. Speed-up is relative to the all-global baseline. Peak memory is essentially constant because, without a backward pass, memory is dominated by weights and image tokens rather than attention activations.\label{tab:local_layer_infer_speed}}
\begin{tabular}{lccc}
\toprule
\#Local layers & Peak mem (GB) & Per-step profile (s) & Speedup \\
\midrule
0 (all global) & 49.65 & 66.76 & 1.00$\times$ \\
\midrule
20 & 49.67 & 47.14 & 1.42$\times$ \\
30 & 49.67 & 37.34 & 1.79$\times$ \\
40 & 49.67 & \textbf{27.53} & \textbf{2.42$\times$} \\
\bottomrule
\end{tabular}
\end{table}

\FloatBarrier

\begin{figure}[htbp]
    \centering
    \includegraphics[width=0.9\linewidth]{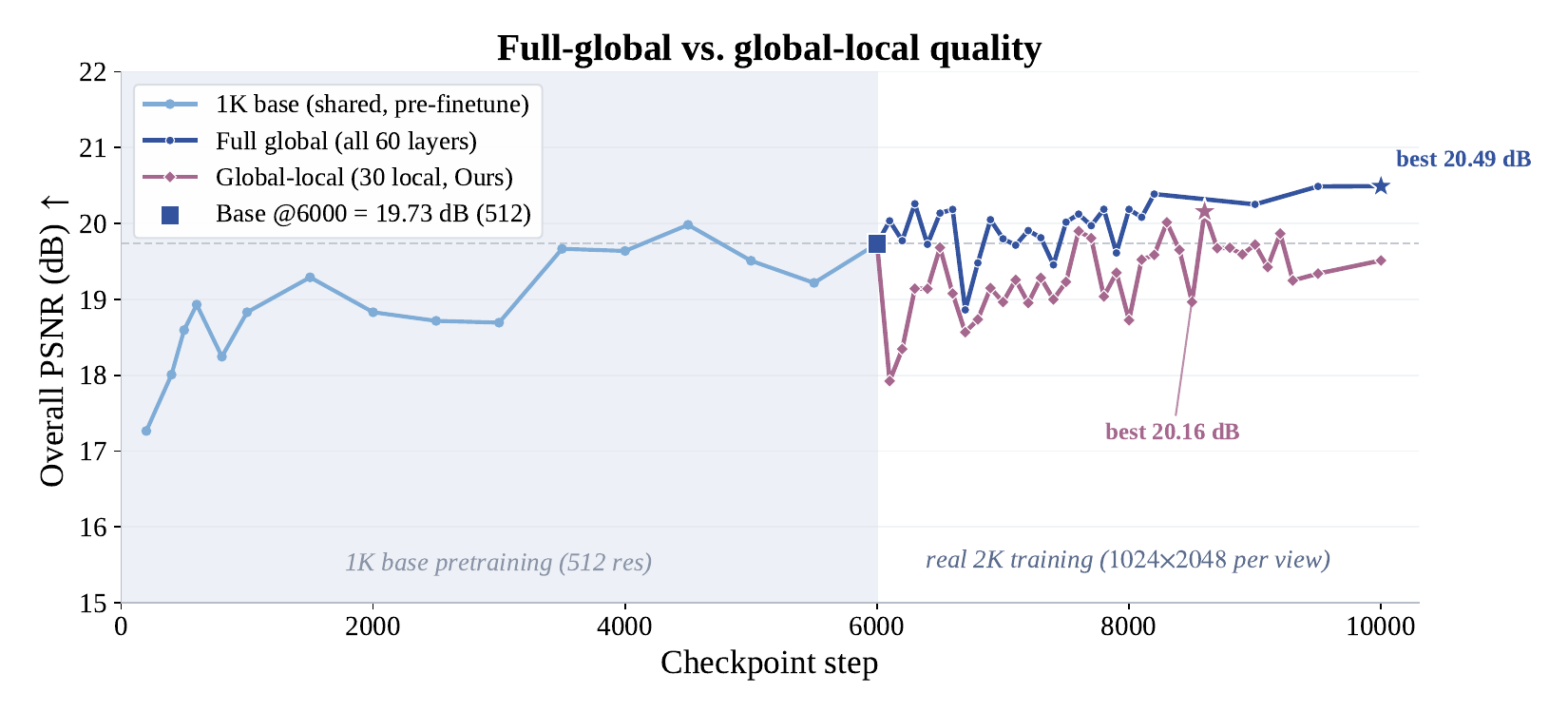}
    \caption{\textbf{Quality side of the global--local trade-off on real 2K.} Overall PSNR versus checkpoint step for full-global attention (all 60 layers) vs.\ our default global--local schedule (30 local layers). The shaded region is the shared 1K base pretraining ($512$ res); at step $6{,}000$ both runs fork from the same base checkpoint (blue square, $19.73$~dB) and are then trained for $2{,}000$ steps at real 2K resolution ($1024{\times}2048$ per view). Each run reports its own best checkpoint ($\star$: $20.49$ vs.\ $20.16$~dB, a $\sim$$0.33$~dB gap), so localizing 30 layers costs little quality while saving substantial memory and time (\Cref{tab:local_layer_tradeoff}). \label{fig:local_global_quality}}
\end{figure}

\noindent\textbf{Is Full-Global Attention Worth Its Cost?} The trade-off above measures compute; here we check the matching quality side under real 2K training. We compare the full-global model (all 60 layers global) against our default global--local schedule (30 least-sensitive layers switched to local), both resumed from the \emph{same} 1K checkpoint at step $6{,}000$. As shown in \Cref{fig:local_global_quality}, the two curves track each other closely: after switching to local attention our schedule quickly rejoins the full-global curve and both fluctuate within a narrow band around the base, with best overall PSNR of $20.49$~dB (full-global) versus $20.16$~dB (global--local)---a gap of only $\sim$$0.33$~dB. \textbf{Full-global attention is thus marginally better in quality but, as quantified in \Cref{tab:local_layer_tradeoff}, substantially more expensive ($90.5$\,GB / $299$\,s per step vs.\ $82.7$\,GB / $169$\,s at 30 local layers). Since the quality drop is small while the compute saving is large, the global--local schedule is a favorable trade-off, which is why we adopt it by default for 2K synthesis.}

\FloatBarrier
\section{Full Per-View NVS Results}
\label{app:per_view_results}
The main paper reports NVS metrics averaged over the four target views for compactness. \Cref{tab:per_view_results_full} preserves the complete per-view results used to compute those averages.

\begin{table}[!ht]
\centering
\caption{\textbf{Full per-view NVS results.} PSNR/SSIM/LPIPS against ground truth at each target pose on UE synthetic and RealSee3D. Best in \textbf{bold}.}
\label{tab:per_view_results_full}
\footnotesize
\setlength{\tabcolsep}{3pt}
\begin{tabular}{ll ccccc ccccc}
\toprule
& & \multicolumn{5}{c}{UE synthetic} & \multicolumn{5}{c}{RealSee3D} \\
\cmidrule(lr){3-7}\cmidrule(lr){8-12}
Method & Metric & V1 & V2 & V3 & V4 & Avg. & V1 & V2 & V3 & V4 & Avg. \\
\midrule
\multirow{3}{*}{GenEx}
& PSNR $\uparrow$ & 15.59 & 15.16 & 15.32 & 14.64 & 15.18 & 15.53 & 15.89 & 16.25 & 15.73 & 15.85 \\
& SSIM $\uparrow$ & 0.477 & 0.465 & 0.467 & 0.437 & 0.461 & 0.638 & 0.643 & 0.654 & 0.642 & 0.644 \\
& LPIPS $\downarrow$ & 0.440 & 0.475 & 0.456 & 0.500 & 0.468 & 0.475 & 0.447 & 0.427 & 0.472 & 0.455 \\
\midrule
\multirow{3}{*}{Matrix-3D}
& PSNR $\uparrow$ & 17.76 & 17.17 & 16.89 & 17.39 & 17.30 & 18.97 & 16.56 & 16.59 & 17.62 & 17.44 \\
& SSIM $\uparrow$ & 0.513 & 0.511 & 0.488 & 0.494 & 0.502 & 0.671 & 0.649 & 0.641 & 0.643 & 0.651 \\
& LPIPS $\downarrow$ & 0.340 & 0.372 & 0.378 & 0.363 & 0.363 & 0.310 & 0.430 & 0.435 & 0.387 & 0.391 \\
\midrule
\multirow{3}{*}{Ours}
& PSNR $\uparrow$ & \textbf{20.61} & \textbf{20.05} & \textbf{19.91} & \textbf{20.06} & \textbf{20.16} & \textbf{19.43} & \textbf{18.14} & \textbf{16.86} & \textbf{17.82} & \textbf{18.06} \\
& SSIM $\uparrow$ & \textbf{0.674} & \textbf{0.663} & \textbf{0.647} & \textbf{0.649} & \textbf{0.658} & \textbf{0.731} & \textbf{0.708} & \textbf{0.702} & \textbf{0.703} & \textbf{0.711} \\
& LPIPS $\downarrow$ & \textbf{0.243} & \textbf{0.256} & \textbf{0.262} & \textbf{0.261} & \textbf{0.256} & \textbf{0.283} & \textbf{0.361} & \textbf{0.400} & \textbf{0.355} & \textbf{0.350} \\
\bottomrule
\end{tabular}
\end{table}

\FloatBarrier
\section{Qualitative Comparisons on UE Synthetic Data}
\label{app:qual}
\begin{figure}[!ht]
    \centering
    \includegraphics[width=1.0\linewidth]{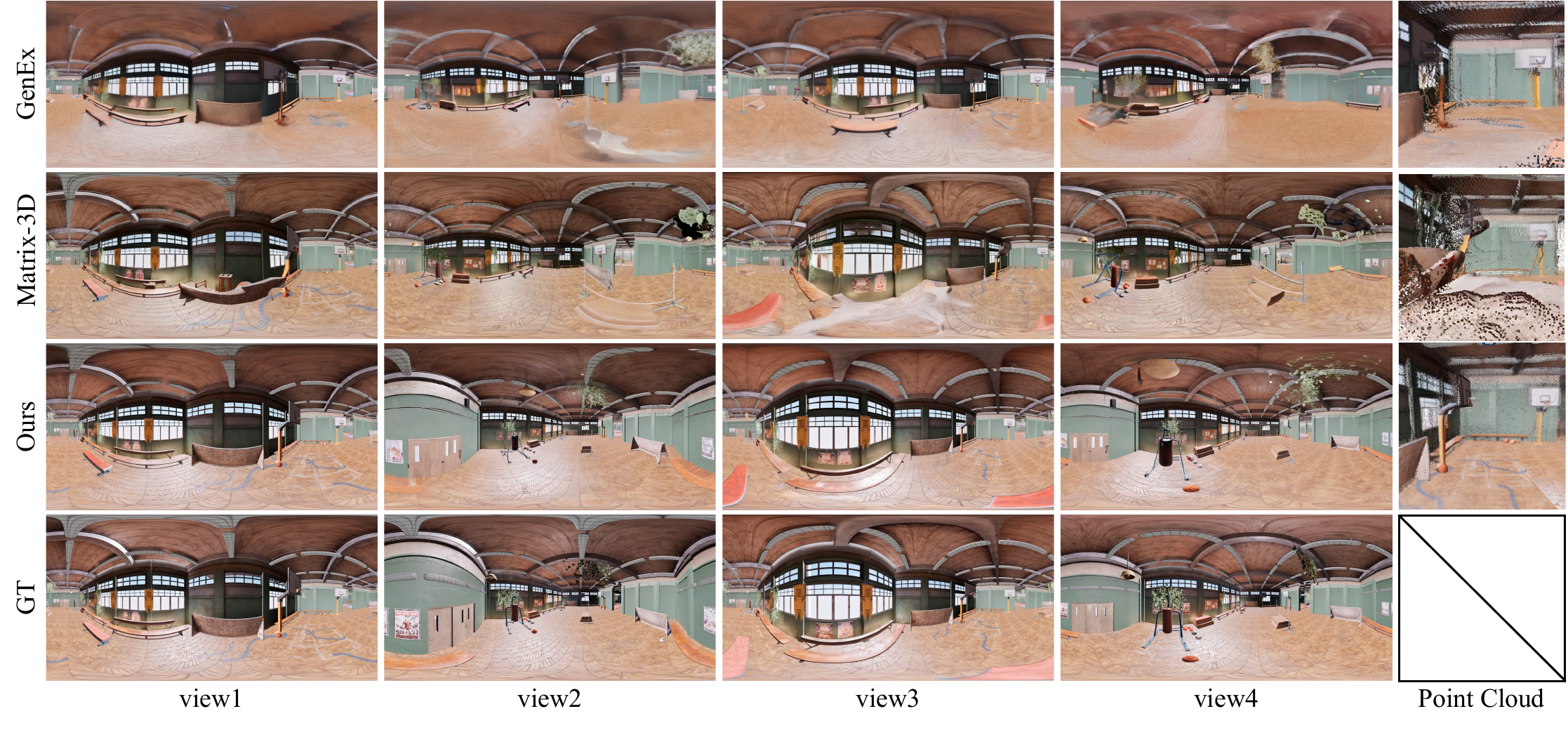}
    \caption{\textbf{Qualitative comparison on a held-out UE synthetic scene.} The four target views generated by WorldExpand preserve scene structure and produce a cleaner downstream point cloud than independently generated panoramic-video keyframes.}
    \label{fig:pano_result_ue}
\end{figure}

\begin{figure}[!ht]
    \centering
    \includegraphics[width=1.0\linewidth]{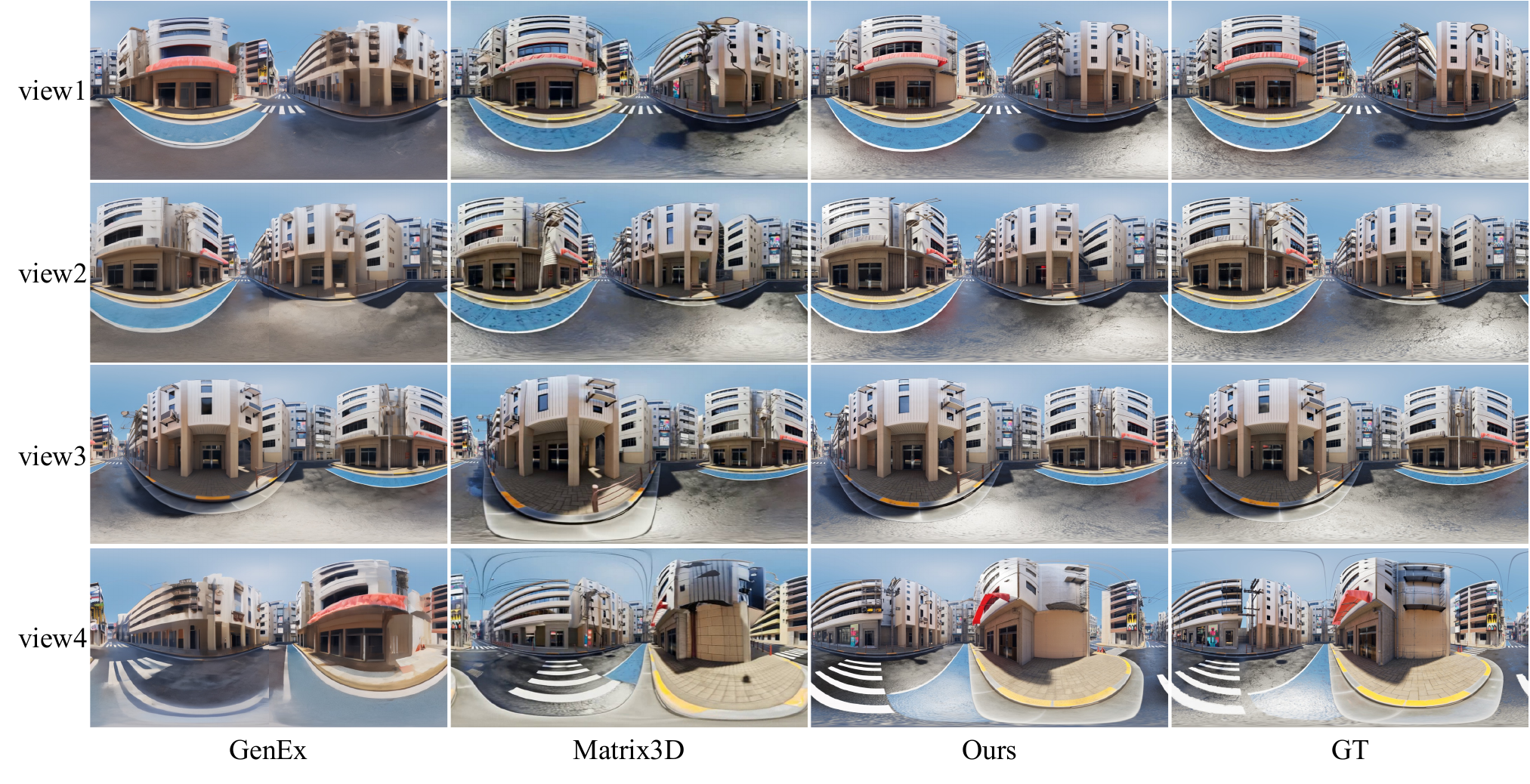}
    \caption{\textbf{Qualitative comparison on novel view synthesis from a single panorama.}
    \label{fig:pano_result2}}
\end{figure}

We provide controlled qualitative comparisons on held-out UE synthetic scenes in \Cref{fig:pano_result_ue,fig:pano_result2}. \textbf{WorldExpand produces sharper structures and more faithful geometry in the target views than the baselines while maintaining consistency with the source panorama.}

\FloatBarrier
\section{Qualitative Comparisons on RealSee3D}
\label{app:realsee}
\begin{figure}[!ht]
    \centering
    \includegraphics[height=0.78\textheight,keepaspectratio]{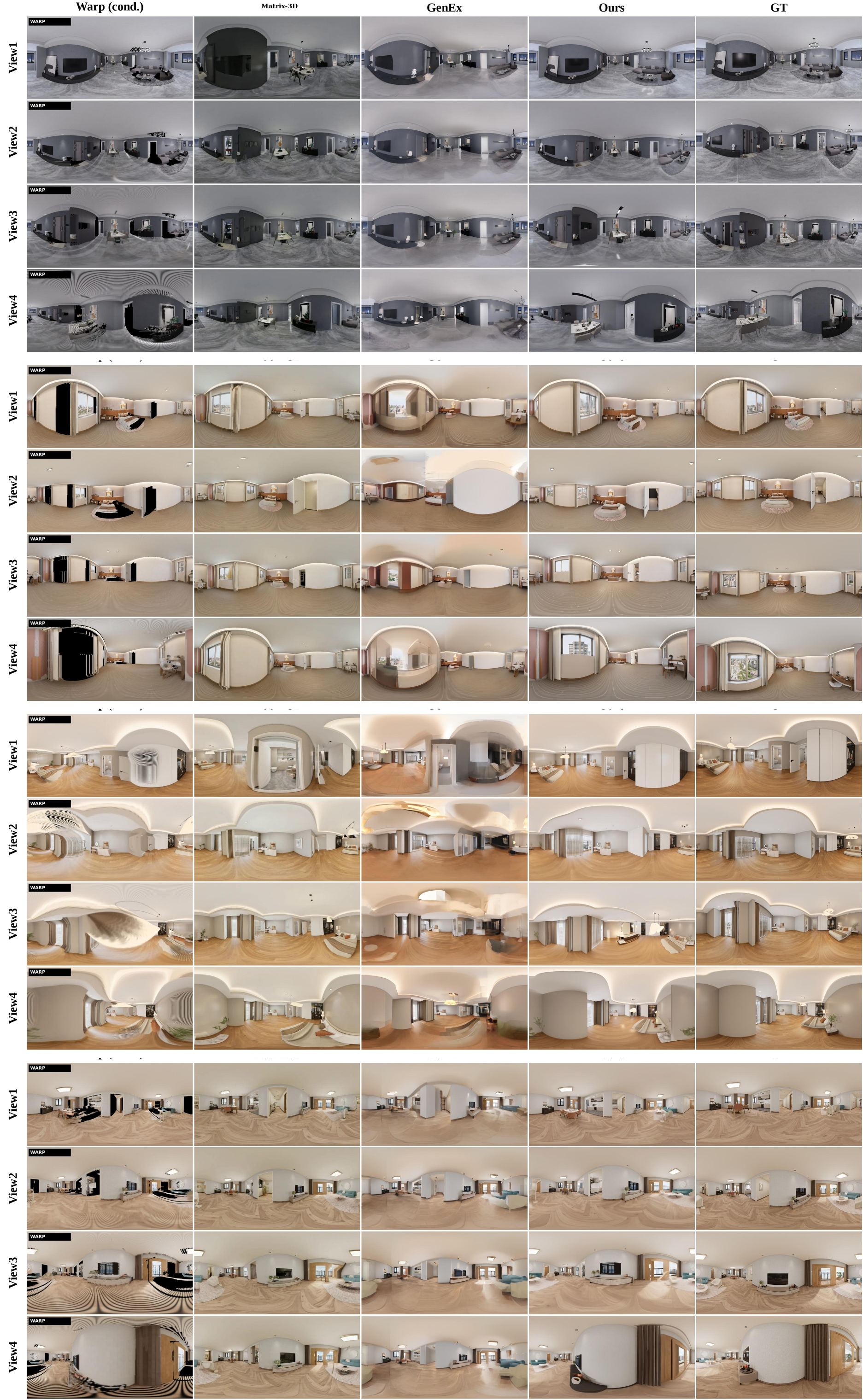}
    \caption{\textbf{Qualitative comparison on RealSee3D.} Four scenes (rows of blocks); within each block we show the four target views (\textsc{View1}--\textsc{View4}). Columns: the point-cloud warp condition (\textsc{Warp}), Matrix-3D, GenEx, our WorldExpand, and the ground-truth panorama (\textsc{GT}). \textbf{Compared with the baselines, WorldExpand synthesizes disoccluded content with sharper structure and more faithful geometry while remaining consistent with the condition and the ground truth.}
    \label{fig:realsee_compare}}
\end{figure}

To complement the quantitative RealSee3D results in the main paper (\Cref{tab:main_results}), \Cref{fig:realsee_compare} shows qualitative novel-view panorama synthesis on the synthetic subset of RealSee3D~\citep{Li2025realsee3d_data}. Despite the domain shift from our UE training data, WorldExpand recovers sharper structures and more faithful geometry in disoccluded regions than the baselines, while remaining consistent with the shared warped condition and ground truth.

\end{document}